\documentclass[11pt]{article}

\usepackage[final]{acl}

\usepackage{times}
\usepackage{latexsym}
\usepackage[T1]{fontenc}
\usepackage[utf8]{inputenc}
\usepackage{microtype}
\usepackage{inconsolata}
\usepackage{graphicx}
\usepackage{booktabs}
\usepackage{amsfonts}
\usepackage{amsmath}
\usepackage{amssymb}
\usepackage{algorithm}
\usepackage{algpseudocode}
\usepackage{multirow}
\usepackage{subcaption}
\usepackage{pifont}
\usepackage{xcolor}
\usepackage{colortbl}
\usepackage{nicefrac}

\definecolor{ourrow}{HTML}{E8F2FF}
\definecolor{bestcell}{HTML}{FFF3CD}

\newcommand{\ours}{DART}
\newcommand{\scroute}{SC-Route}
\newcommand{\think}{\textsc{Think}}
\newcommand{\nothink}{\textsc{NoThink}}

\newcommand{\titleicon}{\raisebox{-0.17\height}{\includegraphics[height=1.25em]{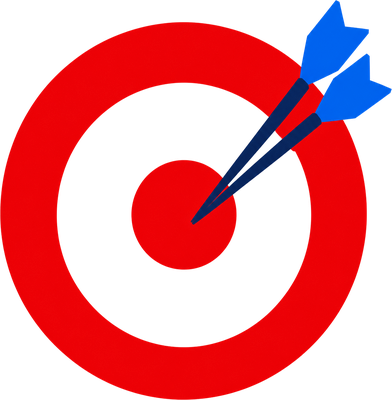}}\hspace{0.28em}}

\title{\titleicon DART: Draft-Agreement Routing for Training-Free \\ Adaptive Thinking Budgets in Hybrid Reasoning Models}

\author{
  \bfseries Jungseob Lee$^{1}$ \enspace
  Seongtae Hong$^{1}$ \enspace
  Seungjun Lee$^{2}$ \enspace
  Jaehyung Seo$^{3}$ \enspace
  Junyoung Son$^{1}$ \\
  \bfseries Sugyeong Eo$^{4}$ \enspace
  Chanjun Park$^{5}$ \enspace
  Hyeongju Park$^{6}$ \enspace
  Hyeonseok Moon$^{1}$\thanks{Corresponding authors.} \enspace
  Heuiseok Lim$^{1}$\footnotemark[1] \\
  $^{1}$Korea University \quad $^{2}$42dot \quad $^{3}$Konkuk University \quad $^{4}$Yonsei University \\
  $^{5}$Soongsil University \quad $^{6}$Kumoh National Institute of Technology \\
  \texttt{\{omanma1928, ghdchlwls123, s0ny, glee889, limhseok\}@korea.ac.kr} \\
  \texttt{seungjun.lee@42dot.ai, seojae777@konkuk.ac.kr, s.eo@yonsei.ac.kr} \\
  \texttt{chanjun.park@ssu.ac.kr, hyungju1203@gmail.com}
}

\begin{document}
\maketitle

\begin{abstract}
Hybrid reasoning models can answer directly or spend extra tokens on extended thinking.
A practical router should choose between these modes for each query, so easy problems avoid unnecessary reasoning and hard problems receive enough budget to finish the answer.
Existing routers move in this direction, but they typically require labeled training data or fix thinking budgets up front, ignoring answer-level evidence from the model itself.
We introduce \ours{}, a training-free routing framework that samples two cheap no-think drafts, accepts direct answering when the drafts agree, and predicts a thinking budget from draft entropy when they disagree.
Across the main comparisons, \ours{} preserves or improves always-thinking accuracy in most settings while reducing thinking-token use.
Accuracy improves by up to $+$9.0 points on Olympiad-level math and by up to $+$22.5 points on code under execution-based equivalence, while thinking-token use drops by 32--73\%.
The Stage~1 signal extends across model scales (0.6B--32B), model families, and API-only hosted settings, with no labeled data and no gradient updates required.
Our code is available at \url{https://github.com/js-lee-AI/DART}.
\end{abstract}

\section{Introduction}
\label{sec:intro}

Recent reasoning-capable large language models (LLMs) increasingly expose hybrid inference interfaces across open-weight models~\citep{qwen3,deepseekr1,deepseekv32,gemma4,yu2025minicpmv45cookingefficient,nvidia2026nemotron3nanoomni} and hosted services~\citep{gpt55,claudethinking}. These systems can invoke an extended-reasoning mode (\think{}), answer directly (\nothink{}), and, in some cases, expose controls over the reasoning budget. The appeal is accuracy and controllability. \think{} mode can substantially improve task accuracy on math, code, and competition benchmarks~\citep{snell2024scaling,efficientreasoning2025}, while budget control lets a deployment assign smaller or larger reasoning budgets according to query difficulty. Together, mode selection and budget allocation let a deployment choose both whether to think and how much reasoning to spend for each query.

However, thinking is not free. For many easy queries the model commits thousands of tokens to deliberate over a problem where a direct response would have sufficed, and recent studies report that extended reasoning sometimes degrades accuracy below the no-think baseline on the easier slice of the same benchmark~\citep{thinkingtrap2025,stopoverthinking2025}. Running every query in \think{} mode is impractical in deployment. Easy queries still pay a substantial thinking cost~\citep{reasoningbudget2025}, while on harder queries the model can exhaust the single-call length limit during the reasoning trace before emitting a complete final answer, unless the deployment provisions a very large token allowance. Provisioning such allowances and the corresponding compute for broad deployment is operationally expensive, making fixed always-thinking deployment difficult to sustain.

Adaptive reasoning routers have emerged to address this overhead by choosing an inference strategy separately for each query, such as direct answering, extended thinking, or assigning a query-specific thinking budget~\citep{selfbudgeter2025,routetoreason2025}. Existing strategies either train a classifier or fine-tune a policy, or rely on confidence and entropy heuristics~\citep{entropycalibration2023,kadavath2022know}. This leaves two practical problems. First, trained routers need labeled difficulty data and model-specific retraining whenever the target model changes, which limits their use in API-only deployments. Heuristic scores avoid retraining but remain indirect proxies for the binary choice between \think{} and \nothink{} mode. Second, common single-call evaluations place the reasoning trace and final answer under one generation cap. Under tight caps, the model can exhaust the budget during reasoning before emitting a complete final answer, so comparing routers against this truncated always-thinking baseline measures a protocol artifact rather than always-thinking accuracy with an intact answer span. Appendix~\ref{app:truncation} quantifies this artifact directly.

In this paper, we introduce \ours{} (\textbf{D}raft-\textbf{A}greement \textbf{R}outing for \textbf{T}hinking), a training-free two-stage router for hybrid reasoning models. \ours{} uses cheap \nothink{} drafts as a query-level difficulty probe. Stage~1 accepts a unanimous draft answer under a pluggable equivalence function and routes only disagreement cases to \think{} mode, while Stage~2 maps draft entropy to a query-specific thinking budget so harder routed queries receive more reasoning and easier routed queries stop earlier. The pipeline requires no labeled difficulty data or gradient updates and separates answer generation from the thinking trace to avoid single-call truncation artifacts. Across the evaluations below, \ours{} reduces thinking-token use without giving up the always-thinking accuracy target.

\textbf{Our contributions are as follows:}
(a) we characterize draft unanimity as a binary difficulty proxy for hybrid reasoning, with a consistently strong correlation to always-thinking correctness.
(b) we propose \ours{}, a training-free router whose Stage~1 draft-agreement decision uses only generated draft answers under a pluggable equivalence function, making it compatible with text-only API access to closed hybrid models, while Stage~2 optionally adds entropy-predicted budgets when token log-probabilities and budget controls are available.
(c) in observed point estimates, \ours{} matches or exceeds always-thinking across Qwen3 8B--32B and DeepSeek-V3.2, improving accuracy on both math and code while using 32--73\% fewer thinking tokens.
(d) the supervised-router diagnostic shows that draft unanimity is more informative than entropy or draft length, reinforcing the Stage~1 signal while \ours{} obtains it without labels or model-specific retraining.

\begin{figure*}[!t]
\centering
\includegraphics[width=\textwidth]{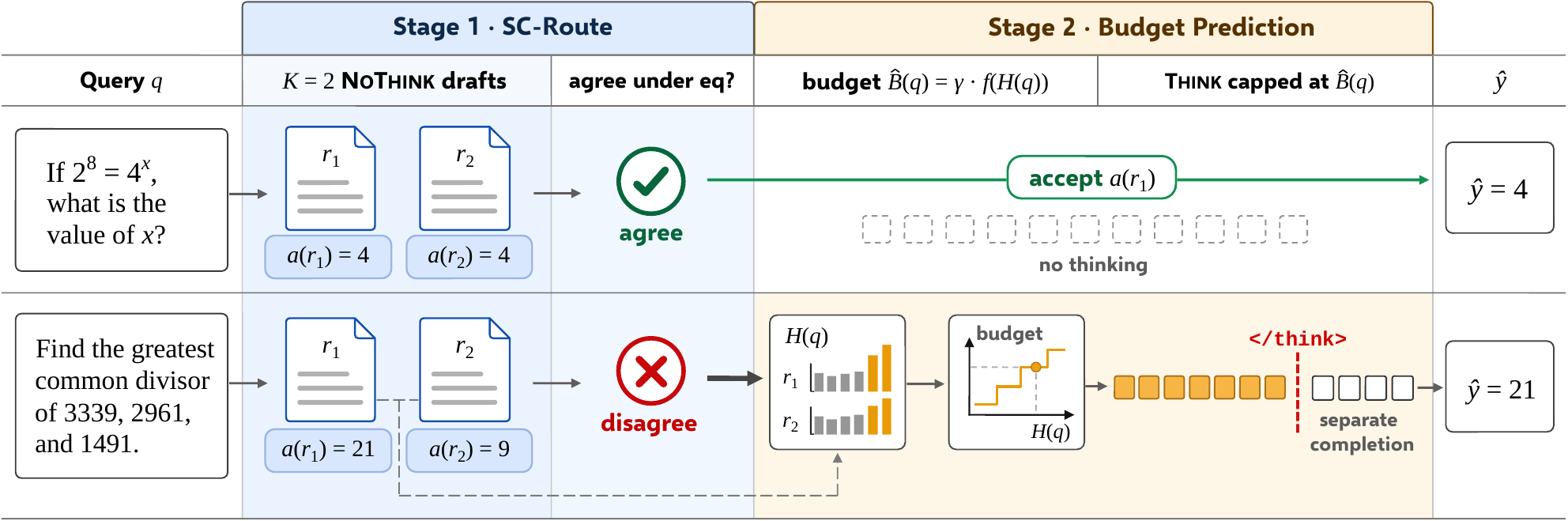}
\caption{Overview of \ours{}. Stage~1 draws $K{=}2$ no-think drafts and accepts the unanimous answer when they agree. On disagreement, Stage~2 maps draft entropy to a query-specific thinking budget and produces the answer in a separate completion.}
\label{fig:framework}
\end{figure*}

\section{Draft-Agreement Routing for Thinking}
\label{sec:method}

\ours{} has two stages. \scroute{} (\S\ref{sec:scroute}) samples $K$ drafts generated in \nothink{} mode and accepts the draft answer only when their extracted answers agree. Stage~2 (\S\ref{sec:scbudget}) allocates a query-specific thinking budget to disagreement cases from draft entropy, concentrating tokens on harder queries. Figure~\ref{fig:framework} gives the pipeline overview and Algorithm~\ref{alg:scroute} the full inference procedure.

\subsection{\texorpdfstring{Preliminaries and Objective}{Preliminaries and Objective}}

Let $q$ be an input query and $\mathcal{M}$ a system exposing two inference modes, \think{} (extended chain-of-thought) and \nothink{} (direct response).
Given a query distribution where some queries admit a correct answer without thinking and others require thinking, our goal is a routing policy $\pi: q \mapsto \{\nothink, \think\}$ that minimizes expected thinking-token cost subject to an accuracy constraint, \emph{without training data or gradient updates}.
This formulation covers both local models exposing a thinking toggle and API-mediated services that route between thinking and direct-answer endpoints.

\subsection{Self-Consistency Routing}
\label{sec:scroute}

Stage~1, \scroute{} (Self-Consistency Routing), adapts the agreement signal used in self-consistency decoding~\citep{selfconsistency2023,aggarwal2023adaptiveconsistency} to routing. When $K$ drafts generated in \nothink{} mode produce the same normalized answer, the model has usually resolved the query without needing an explicit thinking trace.
The routing decision uses two task-facing modules. The answer extractor $a(\cdot)$ maps each draft to the answer object used by the benchmark, and the equivalence function $\text{eq}: \mathcal{A}\!\times\!\mathcal{A}\!\to\!\{0,1\}$ checks whether two extracted answers are equivalent.
The method has one hyperparameter ($K$, the number of drafts) and no thresholds.

\paragraph{Draft.}
Sample $K$ independent completions in \nothink{} mode at temperature $T$.
\begin{equation}
  r_1, \ldots, r_K \;\sim\; \mathcal{M}(q,\; \text{think}{=}\text{false},\; T).
\end{equation}

\paragraph{Route.}
Compute $a(r_k)$ for each draft. If all $K$ extracted answers are pairwise equivalent under $\text{eq}$ (\emph{unanimous agreement}), accept the draft answer. Otherwise, route the query to \think{} mode.
\begin{equation}
  \hat{y} = \begin{cases}
    a(r_1), \quad \text{if drafts agree under eq,} \\
    a(\mathcal{M}(q, \think)), \quad \text{otherwise.}
  \end{cases}
\end{equation}

\paragraph{Design choices.}
We use strict unanimity at $K{=}2$ rather than majority voting because agreement between two independent stochastic samples concentrates probability $\geq p^2$ on the model's dominant answer, while the analysis below shows that this low-cost rule keeps accepted-answer precision high without thresholds and avoids the larger $K$ cost of self-consistency.

\paragraph{Pluggable equivalence.}
\label{sec:equivalence}
The equivalence module $\text{eq}$ is the only domain-specific component. It can be string normalization for math, sandboxed execution match for code, structured-output comparison, and so on. The routing mechanism itself is unchanged across domains. See Appendix~\ref{app:equivalence} for normalization rules and the code execution protocol.

\begin{algorithm}[!t]
\caption{\ours{} Inference Pipeline.}
\label{alg:scroute}
{\footnotesize
\begin{algorithmic}[1]
\Require Query $q$, model $\mathcal{M}$, drafts $K$, temperature $T$
\Ensure Response $\hat{y}$
\State Draw $r_1, \ldots, r_K \sim \mathcal{M}(q,\; \text{think}{=}\text{false},\; T)$
\State $a_i \leftarrow \textsc{ExtractAnswer}(r_i)$ for $i = 1, \ldots, K$
\If{all $a_i$ agree under $\text{eq}$}
  \State \Return $a_1$
\Else
  \State $\hat{B} \leftarrow \gamma \cdot f(H(q))$
  \State $r^* \leftarrow \mathcal{M}(q,\; \text{think}{=}\text{true},\; B{=}\hat{B})$
  \State \Return $\textsc{ExtractAnswer}(r^*)$
\EndIf
\end{algorithmic}
}
\end{algorithm}

\subsection{Budget Prediction and Two-Stage Generation}
\label{sec:scbudget}

Stage~2 is an evaluation-time protocol applied only to queries Stage~1 routes to \think{} mode.

\paragraph{Motivation.}
Disagreement queries vary in how much thinking they need. A fixed budget wastes tokens on the easier slice and truncates the harder slice. We adopt draft entropy as the uncertainty signal for a query-specific budget. Higher draft entropy maps to a larger thinking budget, while lower-entropy queries are served at tight budgets with the final answer produced in a separate completion call. This two-stage design avoids the one-phase measurement artifact where the answer is silently truncated under tight thinking budgets.

Letting $p_t^{(k)}(v) = p_\theta\!\bigl(v \mid q, y^{(k)}_{<t}\bigr)$ denote the model's next-token distribution at position $t$ of draft $k$, draft entropy is the mean token-level entropy over all $K$ drafts.
\begin{equation}
  H(q) = -\frac{1}{K}\!\sum_{k=1}^{K}\!\frac{1}{|\mathcal{T}_k|}\!\sum_{t \in \mathcal{T}_k}\!\sum_{v \in V} p_t^{(k)}(v) \log p_t^{(k)}(v).
  \label{eq:draft_entropy}
\end{equation}
We map entropy to budget via accuracy-label-free isotonic regression~\citep{ayer1955empirical} $f$ on a separate probe run of disagreement-flagged math queries, and apply a safety margin $\gamma{=}1.5$ to keep the predicted budget above the actual thinking-token cost at the matched entropy level.
\begin{equation}
  \hat{B}(q) = \gamma \cdot f\bigl(H(q)\bigr).
  \label{eq:budget_pred}
\end{equation}
For models exposing a stop-thinking control, such as Qwen3, when the thinking trace reaches $\hat{B}(q)$, we inject a \texttt{</think>} token and generate the answer in a separate completion call. The probe run, the isotonic-regression fit, and the choice of $\gamma{=}1.5$ are detailed in Appendix~\ref{app:entropy_budget_fig}.

\definecolor{posdelta}{HTML}{1B5E20}
\definecolor{negdelta}{HTML}{B71C1C}

\begin{table*}[!t]
\centering
\resizebox{0.85\textwidth}{!}{%
\begin{tabular}{@{}ll rrrc rr@{}}
\toprule
& & \multicolumn{4}{c}{\textbf{Accuracy (\%)}} & \multicolumn{2}{c}{\textbf{Efficiency}} \\
\cmidrule(lr){3-6} \cmidrule(lr){7-8}
\textbf{Model} & \textbf{Benchmark} & NT & AT & \scroute{} & \textbf{\ours{}} \footnotesize(vs AT) & Think\,$\downarrow$ & Route\% \\
\midrule
\multirow{4}{*}{Qwen3-8B}
 & MATH-500      & 76.6 & 85.6 & 87.6 & \cellcolor{bestcell}\textbf{88.2}\,{\color{posdelta}\footnotesize($+$2.6)}  & 67\% & 78.0 \\
 & OlympiadBench & 49.8 & \textbf{71.5} & 70.4 & 69.8\,{\color{negdelta}\footnotesize($-$1.7)}  & 45\% & 52.1 \\
 & HumanEval     & 60.4 & 59.1 & 78.7 & \cellcolor{bestcell}\textbf{78.7}\,{\color{posdelta}\footnotesize($+$19.6)} & 55\% & 54.9 \\
 & MBPP          & 59.5 & 64.2 & 67.5 & \cellcolor{bestcell}\textbf{68.9}\,{\color{posdelta}\footnotesize($+$4.7)}  & 46\% & 57.6 \\
\midrule
\multirow{4}{*}{Qwen3-14B}
 & MATH-500      & 81.2 & 87.6 & 87.6 & \cellcolor{bestcell}\textbf{87.6}\,{\color{posdelta}\footnotesize($+$0.0)}  & 73\% & 83.4 \\
 & OlympiadBench & 51.5 & 53.0 & 62.0 & \cellcolor{bestcell}\textbf{62.0}\,{\color{posdelta}\footnotesize($+$9.0)}  & 51\% & 58.0 \\
 & HumanEval     & 71.3 & 66.5 & 78.7 & \cellcolor{bestcell}\textbf{78.7}\,{\color{posdelta}\footnotesize($+$12.2)} & 60\% & 68.9 \\
 & MBPP          & 64.6 & 64.6 & 68.1 & \cellcolor{bestcell}\textbf{68.1}\,{\color{posdelta}\footnotesize($+$3.5)} & 51\% & 63.0 \\
\midrule
\multirow{4}{*}{Qwen3-32B}
 & MATH-500 & 82.2 & 86.2 & 88.5 & \cellcolor{bestcell}\textbf{88.5}\,{\color{posdelta}\footnotesize($+$2.3)} & 69\% & 80.0 \\
 & OlympiadBench & 50.5 & 54.0 & 58.5 & \cellcolor{bestcell}\textbf{58.5}\,{\color{posdelta}\footnotesize($+$4.5)} & 52\% & 59.5 \\
 & HumanEval     & 79.9 & 72.6 & 95.1 & \cellcolor{bestcell}\textbf{95.1}\,{\color{posdelta}\footnotesize($+$22.5)} & 63\% & 76.8 \\
 & MBPP          & 65.8 & 65.8 & 71.2 & \cellcolor{bestcell}\textbf{71.2}\,{\color{posdelta}\footnotesize($+$5.4)} & 51\% & 63.0 \\
\midrule
\multirow{2}{*}{DeepSeek-V3.2$^{\ast}$}
 & MATH-500      & 84.8 & 88.4 & 90.6 & \cellcolor{bestcell}\textbf{92.6}\,{\color{posdelta}\footnotesize($+$4.2)} & 56\% & 85.0 \\
 & OlympiadBench & 63.6 & 66.1 & 67.1 & \cellcolor{bestcell}\textbf{69.1}\,{\color{posdelta}\footnotesize($+$3.0)} & 32\% & 60.1 \\
\bottomrule
\end{tabular}
}
\caption{Accuracy (in \%) and thinking-token efficiency on Qwen3-8B/14B/32B and DeepSeek-V3.2. Accuracy columns report NT, AT, \scroute{} (Stage~1 routing only, with disagreement queries falling back to AT), and the full \ours{} pipeline (with vs-AT delta in \textcolor{posdelta}{green}/\textcolor{negdelta}{red}). Efficiency columns report Think\,$\downarrow$, the thinking-token reduction vs AT, and Route\%, the rate at which Stage~1 accepts unanimous drafts. Shading marks \ours{} matching or exceeding AT, and bold marks the better of \ours{} and AT in each row. $^{\ast}$DeepSeek-V3.2 results use the hosted API.}
\label{tab:main}
\end{table*}

\section{Experimental Setup}
\label{sec:setup}

\subsection{Models}

We evaluate two hybrid reasoning model families, Qwen3 (8B, 14B, 32B)~\citep{qwen3} and DeepSeek-V3.2~\citep{deepseekv32} via the hosted DeepSeek API. The evaluated interfaces expose explicit \think{} and \nothink{} controls through chat-template controls or API endpoints, which we treat as the routing primitive.
The same routing pipeline applies to all backbones with no model-specific tuning, and we use a single shared hyperparameter setting across model families.

\subsection{Benchmarks}

We evaluate on five public benchmarks spanning mathematical reasoning, code generation, and competition math, including
\textbf{MATH-500}~\citep{math500,lightman2023lets} (500 competition-level problems),
\textbf{OlympiadBench}~\citep{olympiadbench} (572 Olympiad-level problems),
\textbf{AIME 2024/2025}~\citep{aime2024,aime2025} (30 problems each),
\textbf{HumanEval}~\citep{humaneval} (164 code generation problems), and
\textbf{MBPP}~\citep{mbpp} (257 code generation problems from the sanitized test split).

\paragraph{Splits.} Code routing executes both drafts against each problem's reference tests, so no code problems or tests are held out and \ours{} rows report all HumanEval and MBPP problems. The Stage~2 entropy-to-budget map is fit label-free on a separate 100-problem probe run, 71 of whose problems also appear in the MATH-500 evaluation set.

\subsection{Baselines and Metrics}

\paragraph{Baselines.}
Our \textbf{always-thinking (AT)} baseline uses a two-phase protocol that emits the thinking trace and the final answer in separate completion calls, eliminating the answer truncation that affects the conventional single-budget one-phase protocol (reported only as a measurement-artifact diagnostic in Appendix~\ref{app:truncation}). \textbf{No-think (NT)} fixes the thinking switch off for all queries. Additional baselines comprise majority voting (MV$k$), paper-defined self-prompted routing baselines (MC-Binary, MC-Conf), and supervised routers (MLP, gradient-boosted trees) trained on labels with draft entropy, draft length, and optional unanimity features.

\paragraph{Metrics.}
We report accuracy, accept precision, and the point-biserial correlation $r_{pb}$ between draft unanimity and AT correctness. Two efficiency metrics complement accuracy. \textbf{Think$\downarrow$} measures the reduction in mean emitted thinking tokens vs.\ AT across all queries, counting actually emitted tokens rather than the predicted budget. \textbf{Route\%} is the unanimous-accept rate, namely the fraction of queries served by Stage~1 with zero thinking tokens.

Evaluation protocol details are given in \S\ref{app:eval_protocol}, and supervised-router CV setup, software versions, sampling temperatures, and hardware in Appendix~\ref{app:implementation}.

\section{Results}
\label{sec:results}

\subsection{Main Results}

Table~\ref{tab:main} summarizes the main comparison across Qwen3 8B, 14B, and 32B on MATH-500, OlympiadBench, HumanEval, and MBPP, together with DeepSeek-V3.2 on MATH-500 and OlympiadBench. Across these observed point estimates, \ours{} reaches the always-thinking accuracy target while reducing emitted thinking tokens on both open-weight Qwen3 models and the hosted DeepSeek-V3.2 API. It matches or exceeds AT on 13 of 14 model--benchmark pairs and reduces thinking-token use by 32--73\%. The single miss is Qwen3-8B on OlympiadBench, where \ours{} trails AT by 1.7 points while still reducing thinking tokens by 45\%. AIME 2024/2025 results are reported as exploratory evidence in Appendix~\ref{app:additional_results}.

\begin{table}[!t]
\centering
\footnotesize
\resizebox{\columnwidth}{!}{%
\begin{tabular}{@{}lccc@{}}
\toprule
\textbf{Method} & \textbf{MATH-500} & \textbf{OlympiadBench} & \textbf{HumanEval} \\
\midrule
\rowcolor{gray!15}
\multicolumn{4}{@{}l}{\textit{Training-free --- Single-mode baselines}} \\
NT & 76.6\,\footnotesize(71\%) & 49.8\,\footnotesize(64\%) & 60.4\,\footnotesize(98\%) \\
AT & 85.6\,\footnotesize(0\%) & \textbf{71.5}\,\footnotesize(0\%) & 59.1\,\footnotesize(0\%) \\
\rowcolor{gray!15}
\multicolumn{4}{@{}l}{\textit{Training-free --- NT aggregation}} \\
MV ($K{=}3$) & 82.4\,\footnotesize(13\%) & 53.7\,\footnotesize($-$7\%) & 10.4\,\footnotesize(95\%) \\
MV ($K{=}5$) & 83.4\,\footnotesize($-$45\%) & 58.6\,\footnotesize($-$79\%) & 10.4\,\footnotesize(91\%) \\
MV ($K{=}7$) & 84.8\,\footnotesize($-$103\%) & 58.7\,\footnotesize($-$151\%) & 9.8\,\footnotesize(88\%) \\
\rowcolor{gray!15}
\multicolumn{4}{@{}l}{\textit{Training-free --- Self-prompted routing}} \\
MC-Binary & 85.8\,\footnotesize(4\%) & 69.9\,\footnotesize(0\%) & 10.4\,\footnotesize($-$7\%) \\
MC-Conf & 85.8\,\footnotesize(24\%) & 54.9\,\footnotesize(25\%) & 8.5\,\footnotesize(44\%) \\
\rowcolor{gray!15}
\multicolumn{4}{@{}l}{\textit{Training-free --- Draft-agreement (ours)}} \\
\textbf{\ours{}} & \textbf{88.2}\,\footnotesize(35\%) & 69.8\,\footnotesize(9\%) & \textbf{78.7}\,\footnotesize(52\%) \\
\rowcolor{gray!15}
\multicolumn{4}{@{}l}{\textit{Supervised routers (5-fold CV on matched benchmark subsets)}} \\
MLP & 81.0\,\footnotesize(69\%) & 58.2\,\footnotesize(64\%) & 62.5\,\footnotesize(98\%) \\
GBT & 78.8\,\footnotesize(64\%) & 54.0\,\footnotesize(51\%) & 63.7\,\footnotesize(88\%) \\
Supv.\ (ent.+len.) & 68.4\,\footnotesize(69\%) & 50.4\,\footnotesize(64\%) & 63.1\,\footnotesize(98\%) \\
Supv.\ +unanimity & 84.0\,\footnotesize(63\%) & 50.0\,\footnotesize(50\%) & 63.1\,\footnotesize(90\%) \\
\bottomrule
\end{tabular}%
}
\caption{Routing strategy comparison on Qwen3-8B. Each cell shows accuracy (\%) and, in parentheses, the mean total-token saving vs AT averaged over all queries (\% reduction in drafts $+$ thinking $+$ answer tokens). Negative values indicate the method generates more total tokens than AT.}
\label{tab:routing}
\end{table}

\begin{figure*}[t]
\centering
\includegraphics[width=0.85\textwidth]{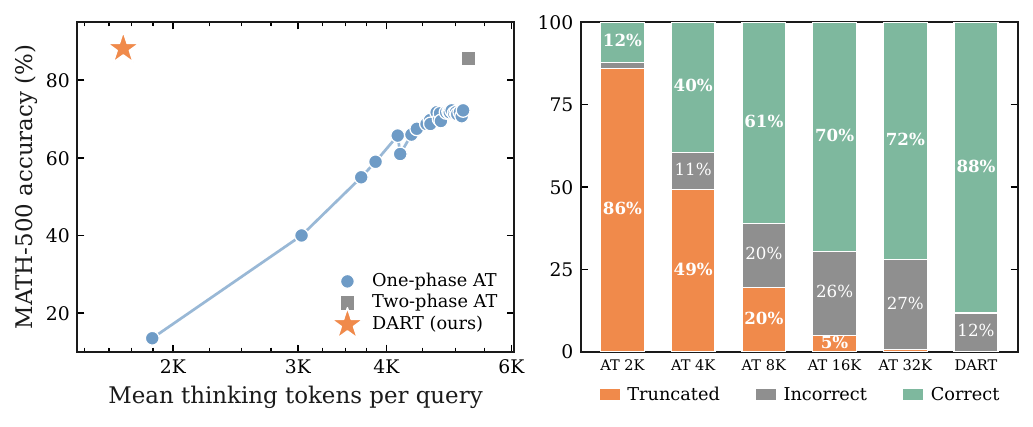}
\caption{Efficiency Pareto (left) and response-level failure-mode breakdown (right) for Qwen3-8B on MATH-500. Both panels place each method at the thinking tokens it actually emits. The one-phase sweep uses the diagnostic grader of \S\ref{app:eval_protocol}, while the two-phase AT and \ours{} points use the main protocol of Table~\ref{tab:main}.}
\label{fig:saturation_and_failure}
\end{figure*}

\paragraph{Math tasks.}
On math, \ours{} improves over AT by up to $+$9.0 points while reducing thinking-token use by 32--73\%. The gains appear on six of eight math model--benchmark pairs, including both DeepSeek-V3.2 rows under the same unanimity rule and without model-specific re-tuning. This transfer matters because DeepSeek-V3.2 is evaluated through a hosted API, so the result tests the router beyond locally hosted Qwen3 checkpoints.

\paragraph{Coding tasks.}
On code, \ours{} improves over AT by up to $+$22.5 points while reducing thinking-token use by 46--63\%. The gains are consistent on HumanEval and MBPP across all three Qwen3 scales. The large HumanEval gains arise because AT underperforms NT on short-form code, and Stage~1 keeps high-agreement code queries on the direct-answer path.

\paragraph{Statistical reliability.}
Table~\ref{tab:signif} reports bootstrap 95\% confidence intervals and McNemar tests computed from the problem-level logs of the Qwen3-8B runs in Table~\ref{tab:main}. The gain of \ours{} over NT is decisive on both benchmarks ($p<.001$). Against AT, \ours{} is statistically indistinguishable, nominally higher on MATH-500 and nominally lower on OlympiadBench, while cutting thinking tokens by 45--67\%, which is the accuracy-at-lower-cost claim the paper makes.

\begin{table}[t]
\centering
\footnotesize
\setlength{\tabcolsep}{2.5pt}
\resizebox{\columnwidth}{!}{%
\begin{tabular}{@{}lcccrr@{}}
\toprule
\textbf{Benchmark} & NT & AT & \textbf{\ours{}} & \multicolumn{2}{c}{McNemar $p$} \\
\cmidrule(l){5-6}
 & \multicolumn{3}{c}{\footnotesize accuracy \tiny[95\% CI]} & \footnotesize vs NT & \footnotesize vs AT \\
\midrule
MATH-500 & 76.6\,\tiny[72.8, 80.2] & 85.6\,\tiny[82.4, 88.6] & \textbf{88.2}\,\tiny[85.2, 91.0] & $<.001$ & .066 \\
OlympiadBench & 49.8\,\tiny[45.8, 53.8] & \textbf{71.5}\,\tiny[67.7, 75.2] & 69.8\,\tiny[65.9, 73.4] & $<.001$ & .30 \\
\bottomrule
\end{tabular}%
}
\caption{Bootstrap 95\% confidence intervals and McNemar tests for Qwen3-8B, computed from the main runs.}
\label{tab:signif}
\end{table}

\subsection{Routing Strategy Comparison}

Table~\ref{tab:routing} compares \ours{} against training-free and supervised baselines on Qwen3-8B. MV$k$ samples multiple \nothink{} drafts and returns a majority answer, so it aggregates direct answers rather than deciding when to think. MC-Binary asks the model whether thinking is needed, and MC-Conf routes from a self-reported confidence score. The supervised rows train MLP or gradient-boosted tree routers with 5-fold cross-validation, using draft entropy, mean draft length, and, where indicated, Stage-1 unanimity. HumanEval supervised rows use execution-based labels, and \ours{} HumanEval is reported on the full benchmark.

\paragraph{Training-free baselines.}
MV$k$ ($K{=}3$--$7$) plateaus below AT on math and collapses on HumanEval to below 11\%. The self-prompted MC-Binary row reaches 85.8\% on MATH-500 but also collapses on HumanEval to 10.4\%.
Token-level vote aggregation and self-prompted confidence both fail on code where draft-level execution agreement succeeds, suggesting agreement captures a domain-agnostic difficulty signal the others miss. The MV$k$ rows instantiate \nothink{} parallel scaling~\citep{ma2025nothinking}, whose vote saturates at 84.8 by $K{=}7$ on MATH-500. Difficulty-Adaptive Self-Consistency~\citep{wang2024penny}, instantiated over the same drafts with a confidence-based stopping rule, returns the identical majority answer on all 500 MATH-500 problems while drawing 4.9 samples on average instead of seven, so it lowers sampling cost without lifting that ceiling. Neither verifier-free aggregation reaches \ours{}'s 88.2, because majority voting over \nothink{} drafts never makes the think-or-not decision.

\paragraph{Supervised routers.}
\label{sec:supervised}
Even with 250 labels, gradient-boosted trees (78.8\%) and MLPs (81.0\%) underperform training-free \ours{} (88.2\% on MATH-500).
A supervised router augmented with the draft-unanimity feature improves to 84.0\%, but still trails training-free \ours{}. Its learned unanimity coefficient ($-4.43$) is much larger than entropy's ($-0.13$), showing that the diagnostic supervised model rediscovers the same signal that \ours{} uses directly without labels.
Without the unanimity feature, supervised routers learn a degenerate ``always NT'' policy.

\subsection{Token and Latency Efficiency}
\label{sec:token_budget}

\paragraph{Token budget.}
\ours{} reduces generated tokens while improving accuracy over AT. Table~\ref{tab:think_budget} reports mean generation tokens on Qwen3-8B against the same two-phase AT baseline as Table~\ref{tab:main}, counting drafts, thinking traces, and answers on MATH-500 and thinking traces alone on HumanEval.

\begin{table}[t]
\centering
\footnotesize
\resizebox{0.9\columnwidth}{!}{%
\begin{tabular}{@{}lrrrc@{}}
\toprule
\textbf{Benchmark} & \textbf{AT tok.} & \textbf{\ours{} tok.} & \textbf{Savings} & $\Delta$\,Acc \\
\midrule
MATH-500       & 5{,}382  & 3{,}475  & {\color{posdelta}$-$35\%} & {\color{posdelta}$+$2.6} \\
HumanEval & 2{,}980 & 1{,}339 & {\color{posdelta}$-$55\%} & {\color{posdelta}$+$19.6} \\
\bottomrule
\end{tabular}%
}
\caption{Mean generation tokens and accuracy delta against the two-phase AT baseline on Qwen3-8B. The MATH-500 row counts drafts, thinking, and answers, the HumanEval row thinking only.}
\label{tab:think_budget}
\end{table}

On MATH-500, \ours{} averages 3.5K total generated tokens compared with 5.4K for AT and gains 2.6 accuracy points, yielding a 35\% total-token reduction at positive accuracy delta. On HumanEval, \ours{} reduces thinking tokens by 55\% while gaining 19.6 points. The code saving arises because accepted \nothink{} drafts are short code snippets, whereas AT spends long reasoning traces on many queries that Stage~1 answers directly. Figure~\ref{fig:saturation_and_failure} complements the table for MATH-500 by visualizing the efficiency frontier and separating false accepts from \think{}-mode failures for the error analysis in Section~\ref{sec:error_analysis}.

\paragraph{Wall-clock latency.}
\label{sec:latency}
Token savings also reduce wall-clock latency. Table~\ref{tab:latency} in Appendix~\ref{app:latency} reports that \ours{} averages 29.5 seconds on Qwen3-8B MATH-500, compared with 67.6 seconds for AT. The 2.3-fold speedup comes from the 78\% of queries accepted after parallel \nothink{} drafts, which finish in 14.6 seconds on average and avoid a \think{} pass.

\section{Analysis}
\label{sec:analysis}

\subsection{\texorpdfstring{Difficulty Alignment and Generalization}{Difficulty Alignment and Generalization}}

This analysis tests the assumption behind Stage~1. Draft agreement should mark queries that can be answered directly, while disagreement should identify cases that need \think{} mode. Under this assumption, Stage~1 should accept fewer examples as benchmark difficulty rises while keeping accepted-answer precision high. The signal should also remain positive across model families and scales.

\begin{figure}[t]
\centering
\includegraphics[width=\columnwidth]{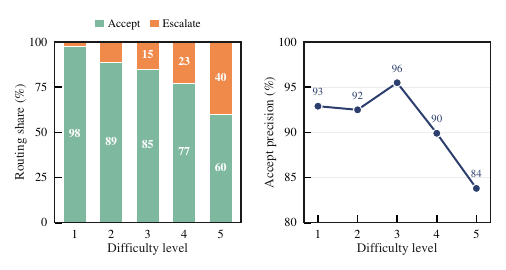}
\caption{Difficulty-stratified Stage-1 behavior on MATH-500 (Qwen3-8B). The left panel shows the no-think accept share and the share escalated to thinking across difficulty levels 1--5. The right panel shows precision among accepted no-think answers at each level.}
\label{fig:difficulty_routing}
\end{figure}

\paragraph{Difficulty alignment.}
Stage~1 accepts fewer queries as annotated difficulty increases while preserving high precision among accepted answers. Figure~\ref{fig:difficulty_routing} reports MATH-500 difficulty levels for Qwen3-8B. The Stage~1 accept rate decreases monotonically from 97.7\% at Level~1 to 59.7\% at Level~5, while accepted-answer precision remains between 83.8\% and 95.5\% across all five levels. This pattern supports using agreement as the Stage~1 gate because it identifies the subset where direct \nothink{} answers are reliable and sends the remaining harder problems to \think{} mode. Aggregated over all queries, accepted answers are 90.8\% correct on MATH-500 and 81.9\% on OlympiadBench, 14 and 32 points above the unconditional \nothink{} accuracies of 76.6 and 49.8 in Table~\ref{tab:main}, so the accept-precision-versus-base-rate gap measures the gate's strength directly.

\paragraph{Robustness across model families.}
Using the model--benchmark settings summarized in Table~\ref{tab:main} and the Qwen3 size sweep in Table~\ref{tab:scale}, draft agreement remains positively associated with AT correctness across every tested pair. The point-biserial correlation $r_{pb}$ is positive and significant ($p < 0.001$) across two model lineages, two architectures, and scales from 0.6B to 32B. The Qwen3 scale comparison is stable, with $r_{pb} = 0.56$ at 8B and $0.57$ at 32B, indicating that the signal is not confined to one parameter scale.

\paragraph{Scale transfer.}
Token savings persist across the Qwen3 scale sweep, but accuracy parity depends on model capability. Table~\ref{tab:scale} reports Qwen3 0.6B--32B results for the benchmark rows completed under the matched full-pipeline protocol. \ours{} trails AT by 2.0--7.5 points on MATH-500 at 0.6B--4B. At 8B--32B, it matches or exceeds AT on all scale rows except Qwen3-8B OlympiadBench, the same $-$1.7 point exception reported in the main comparison, while retaining 45--73\% thinking-token reductions. The scale sweep explains when the routing signal is reliable enough to match AT.

\begin{table}[t]
\centering
\small
\setlength{\tabcolsep}{3pt}
\resizebox{\columnwidth}{!}{%
\begin{tabular}{@{}ll rrr r@{}}
\toprule
\textbf{Model} & \textbf{BM} & \textbf{NT} & \textbf{AT} & \textbf{\ours{}} \footnotesize(vs AT) & Think\,$\downarrow$ \\
\midrule
\multirow{3}{*}{0.6B}
 & MATH-500  & 38.5 & \textbf{56.0} & 54.0\,{\color{negdelta}\footnotesize($-$2.0)} & 40\% \\
 & HumanEval & 27.4 & 48.8 & \textbf{50.6}\,{\color{posdelta}\footnotesize($+$1.8)} & 9\% \\
 & MBPP      & 34.6 & 48.2 & \textbf{51.4}\,{\color{posdelta}\footnotesize($+$3.2)} & 16\% \\
\midrule
\multirow{3}{*}{1.7B}
 & MATH-500  & 52.5 & \textbf{65.5} & 58.0\,{\color{negdelta}\footnotesize($-$7.5)} & 47\% \\
 & HumanEval & 55.5 & 85.4 & \textbf{86.6}\,{\color{posdelta}\footnotesize($+$1.2)} & 40\% \\
 & MBPP      & 48.2 & 57.2 & \textbf{61.9}\,{\color{posdelta}\footnotesize($+$4.7)} & 28\% \\
\midrule
\multirow{3}{*}{4B}
 & MATH-500  & 66.5 & \textbf{70.5} & 65.5\,{\color{negdelta}\footnotesize($-$5.0)} & 48\% \\
 & HumanEval & 67.1 & 74.4 & \textbf{91.5}\,{\color{posdelta}\footnotesize($+$17.1)} & 58\% \\
 & MBPP      & 56.4 & 62.6 & \textbf{66.9}\,{\color{posdelta}\footnotesize($+$4.3)} & 41\% \\
\midrule
\multirow{4}{*}{8B}
 & MATH-500  & 76.6 & 85.6 & \textbf{88.2}\,{\color{posdelta}\footnotesize($+$2.6)} & 67\% \\
 & OlympiadBench & 49.8 & \textbf{71.5} & 69.8\,{\color{negdelta}\footnotesize($-$1.7)} & 45\% \\
 & HumanEval & 60.4 & 59.1 & \textbf{78.7}\,{\color{posdelta}\footnotesize($+$19.6)} & 55\% \\
 & MBPP      & 59.5 & 64.2 & \textbf{68.9}\,{\color{posdelta}\footnotesize($+$4.7)} & 46\% \\
\midrule
\multirow{4}{*}{14B}
 & MATH-500   & 81.2 & 87.6 & \textbf{87.6}\,{\footnotesize($+$0.0)} & 73\% \\
 & OlympiadBench & 51.5 & 53.0 & \textbf{62.0}\,{\color{posdelta}\footnotesize($+$9.0)} & 51\% \\
 & HumanEval & 71.3 & 66.5 & \textbf{78.7}\,{\color{posdelta}\footnotesize($+$12.2)} & 60\% \\
 & MBPP      & 64.6 & 64.6 & \textbf{68.1}\,{\color{posdelta}\footnotesize($+$3.5)} & 51\% \\
\midrule
\multirow{4}{*}{32B}
 & MATH-500      & 82.2 & 86.2 & \textbf{88.5}\,{\color{posdelta}\footnotesize($+$2.3)} & 69\% \\
 & OlympiadBench & 50.5 & 54.0 & \textbf{58.5}\,{\color{posdelta}\footnotesize($+$4.5)} & 52\% \\
 & HumanEval     & 79.9 & 72.6 & \textbf{95.1}\,{\color{posdelta}\footnotesize($+$22.5)} & 63\% \\
 & MBPP          & 65.8 & 65.8 & \textbf{71.2}\,{\color{posdelta}\footnotesize($+$5.4)} & 51\% \\
\bottomrule
\end{tabular}%
}
\caption{\ours{} across Qwen3 model sizes (0.6B--32B).}
\label{tab:scale}

\end{table}

\paragraph{Routing hyperparameter sensitivity.}
Table~\ref{tab:k3} and Appendix~\ref{app:hparam_ablation} report the $K$ sweep, and Appendix~\ref{app:budget} reports thinking-budget cap and draft-temperature sweeps. Unanimity at $K{=}2$ trades accept rate against accept precision more favorably than $K{=}3$ majority voting. The budget sweep shows that accuracy plateaus once the cap exceeds the entropy-predicted average, and temperature sensitivity is mild around the chat-template defaults ($T{=}0.6$--$0.8$). Higher $T$ inflates draft disagreement without an accuracy benefit. Appendix~\ref{app:self_verify} reports a self-verification alternative that detects only 22.2\% of wrong \nothink{} drafts.

Table~\ref{tab:k3} contrasts unanimity at $K{=}2$ and $K{=}3$ with $K{=}3$ majority voting on MATH-500. Unanimity trades accept rate for accept precision, while majority voting at $K{=}3$ accepts every query at substantially lower precision.

\begin{table}[t]
\centering
\small
\begin{tabular}{@{}lccc@{}}
\toprule
$K$ & Decision rule & Accept\% & Prec.\% \\
\midrule
2 & Unanimity & 78.0 & 90.8 \\
3 & Unanimity & 73.6 & 93.5 \\
3 & Majority vote & 100.0 & 82.4 \\
\bottomrule
\end{tabular}
\caption{Effect of the number of drafts $K$ on routing quality (Qwen3-8B, MATH-500).}
\label{tab:k3}

\end{table}

Table~\ref{tab:mcq} in Appendix~\ref{app:hparam_ablation} reports the routing signal on three multiple-choice benchmarks (ARC-Challenge~\citep{arc}, MMLU-Pro~\citep{mmlupro}, GPQA-D~\citep{gpqa}), where the point-biserial correlation between unanimity and AT correctness weakens sharply (0.13, 0.06, $-$0.009), unlike the open-ended math/code settings of Table~\ref{tab:main}. The failure has a structural explanation. For $K{=}2$ drafts from distribution $\mathbf{p} = (p_1, \ldots, p_C)$ over $C$ options, the collision probability is $P(\text{agree}) = \sum_{i=1}^{C} p_i^2$. For a 4-choice MCQ where the model places 60\% mass on one option, $P(\text{agree}) \geq 0.36$, so agreement occurs frequently even when the model is wrong, breaking the unanimity precision required by Stage~1.

\subsection{Error Analysis}
\label{sec:error_analysis}

\paragraph{Oracle routing analysis.}
For the Qwen3-8B MATH-500 setting in Table~\ref{tab:main}, an oracle routing diagnostic that selects the better mode for each query with ground-truth labels reaches 91.2\%.
\ours{} reaches 88.2\% in the same setting, capturing 79\% of the oracle gain over the 76.6\% no-think baseline.
The remaining 3.0-point gap motivates the response-level error taxonomy below, visualized in the right panel of Figure~\ref{fig:saturation_and_failure}.

\paragraph{Error taxonomy.}
Of \ours{}'s 59 errors on MATH-500, 36 are false accepts (61\%), cases where drafts agree on the wrong answer, and 23 are \think{}-mode failures (39\%).
False accepts are characterized by 1.38-fold higher draft entropy and 1.71-fold longer drafts compared to true accepts, suggesting these queries are challenging but not separated by the current unanimity gate.
This analysis points to the clearest future improvement, reducing false accepts through confidence calibration or additional draft rounds.

\subsection{Measurement Artifacts and Fair Comparison}

\paragraph{The one-phase accuracy gap is mostly truncation.}
Table~\ref{tab:ts_vs_op} compares two-stage and one-phase measurement at matched budgets on MATH-500, where the two-stage column runs \ours{} at a fixed budget. One-phase AT reaches only 13.5\% at 2K with 86.0\% of responses truncated and recovers to 72.0\% once truncation falls below 1\%, so its accuracy tracks its own truncation rate. We therefore adopt the two-phase protocol for the AT baseline. A broader budget sweep from 2K to 60K is provided in Appendix~\ref{app:truncation}.

\paragraph{Matched-budget accuracy depends on the measurement protocol.}
Table~\ref{tab:ts_vs_op} reports this comparison at eight budget points from 2K to 32K. The accuracy gap $\Delta$ collapses monotonically from 56 points at 2K to 0.25 points at 32K, tracking the one-phase truncation rate almost exactly (Pearson $r{>}0.99$). Thinking capacity is not matched across the two columns, since the one-phase column emits 2.3-fold to 4.4-fold more thinking tokens at every budget. The gap therefore has two sources this table cannot separate, an intact answer span and the 78 to 82 percent of queries Stage~1 answers directly. It does establish that one-phase measurement understates a fixed-budget baseline most severely at tight budgets.

\begin{table}[t]
\centering
\small
\resizebox{0.99\columnwidth}{!}{%
\begin{tabular}{@{}r rrr r@{}}
\toprule
\textbf{Budget} & \textbf{2-stage} & \textbf{1-phase} & $\Delta$ & \textbf{Trunc\%} \\
\midrule
 2K & 69.75 & 13.50 & {\color{posdelta}$+$56.25} & 86.0 \\
 4K & 70.00 & 40.00 & {\color{posdelta}$+$30.00} & 49.2 \\
 6K & 71.25 & 55.00 & {\color{posdelta}$+$16.25} & 28.7 \\
 8K & 70.00 & 61.00 & {\color{posdelta}$+$9.00}  & 19.5 \\
12K & 71.75 & 68.75 & {\color{posdelta}$+$3.00}  &  9.2 \\
16K & 72.25 & 69.50 & {\color{posdelta}$+$2.75}  &  5.0 \\
24K & 72.25 & 70.75 & {\color{posdelta}$+$1.50}  &  2.8 \\
32K & 72.25 & 72.00 & {\color{posdelta}$+$0.25}  &  0.8 \\
\midrule
about 2.8K (\textbf{\ours{}}) & \textbf{88.2} & \multicolumn{1}{c}{---} & \multicolumn{1}{c}{---} & 0.0 \\
\bottomrule
\end{tabular}%
}
\caption{Matched-budget measurement protocols on MATH-500 (Qwen3-8B). Two-stage emits thinking and answer in separate completions, while one-phase combines them in one budget. Budget rows run \ours{} at a fixed budget under a string-extract grader on 400 problems. The final row is the main-protocol result of Table~\ref{tab:main}. Trunc\% the one-phase truncation rate.}
\label{tab:ts_vs_op}
\end{table}

\paragraph{One-phase AT plateaus under increased budget.}
The extended one-phase budget sweep in Appendix~\ref{app:truncation} shows accuracy plateaus near 72\% from 28K onward, with truncation rate at or below 1\%. Under the main evaluation protocol, two-stage AT reaches 85.6\% and \ours{} reaches 88.2\% while emitting 1{,}743 mean thinking tokens against AT's 5{,}343, the 3.1-fold reduction in Table~\ref{tab:iso_accuracy}. No fixed one-phase budget matches \ours{}'s accuracy under the matched comparison. The same truncation collapse and high-budget plateau replicate on OlympiadBench, showing the pattern on a second math benchmark.

\begin{table}[t]
\centering
\footnotesize
\setlength{\tabcolsep}{4pt}
\begin{tabular}{@{}l c r r c@{}}
\toprule
\textbf{Method} & \textbf{Budget} & \textbf{Acc} & \textbf{Tokens} & \textbf{Reduction} \\
\midrule
AT two-stage & 16K & 85.6 & 5{,}343 & 1.0$\times$ \\
NT no-think  & --- & 76.6 & 0 & --- \\
\textbf{\ours{}} & about 2.8K & \textbf{88.2} & \textbf{1{,}743} & \textbf{3.1$\times$} \\
\bottomrule
\end{tabular}
\caption{Iso-accuracy comparison on MATH-500 (Qwen3-8B). Budget is each method's thinking-token allowance, a 16k-token cap for AT and a per-query entropy-predicted value for \ours{}. Tokens is the mean emitted thinking-token count over all 500 queries, and reduction is AT-tokens / \ours{}-tokens. Only 24 of 500 AT queries reach the cap.}
\label{tab:iso_accuracy}
\end{table}

\section{Related Work}
\label{sec:related}

\paragraph{Adaptive reasoning and test-time compute.}
Recent work studies how to allocate test-time computation for reasoning models. Some approaches learn routing or control policies for reasoning depth, including AdaptThink~\citep{adaptthink2025}, ThinkSwitcher~\citep{thinkswitcher2025}, HBPO~\citep{hbpo2025}, and Route to Reason~\citep{routetoreason2025}. Others predict the amount of reasoning to spend, such as SelfBudgeter~\citep{selfbudgeter2025}. Recent surveys~\citep{snell2024scaling,efficientreasoning2025,reasoningbudget2025} organize this broader landscape, while related studies examine hybrid-thinking controllability~\citep{demystifying2025}, over-thinking~\citep{thinkingtrap2025,stopoverthinking2025,lee2026acoer}, and adaptive mode switching~\citep{asrr2025}. Table~\ref{tab:router_comparison} summarizes deployment constraints for the router-style methods most closely related to our setting.

\begin{table}[t]
\centering
\footnotesize
\setlength{\tabcolsep}{2pt}
\begin{tabular}{@{}lcccc@{}}
\toprule
Method & Train & Logits & API & Retune \\
\midrule
AdaptThink & GRPO & No & \ding{55} & Retrain \\
ThinkSw. & Supervised & No & \ding{55} & Retrain \\
SelfBudg. & Supervised & Yes & \ding{55} & Retrain \\
HBPO & RL & No & \ding{55} & Retrain \\
R2Reason & RL & No & \ding{55} & Retrain \\
\midrule
\textbf{\ours{}} & \textbf{None} & Stage~2$^*$ & Stage~1 & \textbf{None} \\
\bottomrule
\end{tabular}
\caption{Training requirements and deployment constraints of adaptive reasoning routers. $^*$Used only for Stage~2 budget prediction.}
\label{tab:router_comparison}
\end{table}

\paragraph{Self-consistency and confidence proxies.}
Self-consistency~\citep{selfconsistency2023} samples multiple chains and aggregates by majority vote. Adaptive Consistency~\citep{aggarwal2023adaptiveconsistency} dynamically adjusts the sample count based on running agreement. Semantic entropy~\citep{entropycalibration2023} clusters paraphrastically equivalent answers and treats the resulting distribution as a confidence proxy. Universal Self-Consistency~\citep{universalsc2024} extends this idea to free-form generation through an LLM-judged equivalence test. These methods establish agreement and answer dispersion as useful signals for generation-time uncertainty~\citep{lee2026pricecheck,lee2026directional}.

\paragraph{Routing across models or computation paths.}
Routing has also been studied outside hybrid reasoning~\citep{lee2026madara,lee2026redline}. FrugalGPT~\citep{frugalgpt2023} and RouteLLM~\citep{routellm2024} route queries across separate models of different cost. Speculative decoding~\citep{leviathan2023speculativedecoding,lee2026offpolicy,lee2026cast,lee2026glance} coordinates token-level generation between a drafter and a verifier. Confident Adaptive Language Modeling~\citep{schuster2022calm} exits early from decoding when confidence is high, and Adaptive Computation Time~\citep{graves2016act} learns halting decisions in recurrent networks.

\section{Conclusion}
We introduced \ours{}, a training-free router for hybrid reasoning models that uses agreement among cheap \nothink{} drafts as answer-level evidence for when direct answering is sufficient. When drafts disagree, \ours{} routes the query to \think{} mode and can allocate an entropy-predicted thinking budget, separating the reasoning trace from final-answer generation to avoid truncation artifacts. Across the main comparisons, \ours{} matches or exceeds always-thinking across benchmarks while reducing thinking-token use by 32--73\%, with the largest accuracy gains on code under execution-based equivalence. The analysis shows that draft agreement tracks difficulty across model families and scales (0.6B--32B), and it identifies false accepts as the main remaining error source, pointing to lightweight calibration or selective extra drafts as natural next steps.

\section*{Limitations}

\paragraph{Scope and residual errors.}
\ours{} targets open-ended generation with a large answer space, such as mathematical reasoning and code generation. Multiple-choice tasks fall outside this scope because a small option set makes chance agreement likely even when the shared answer is wrong. Table~\ref{tab:mcq} in Appendix~\ref{app:hparam_ablation} reports that the point-biserial correlation between draft unanimity and AT correctness weakens sharply there, so we exclude them rather than retrofit a confidence threshold. Even within scope, draft agreement is a high-precision signal rather than a correctness guarantee. When two \nothink{} drafts converge on the same wrong answer, \ours{} accepts without invoking \think{} mode, and Section~\ref{sec:error_analysis} identifies these false accepts as the main residual error source on MATH-500.

\paragraph{Equivalence functions.}
The routing decision depends on a domain-specific, pluggable answer extractor and answer-equivalence function, using answer normalization for math and sandboxed execution against each problem's reference tests for code. Each new domain needs its own equivalence check, neither too permissive nor too brittle, and a weak check can inflate false accepts or unnecessarily reduce accept rates. For code the check is not independent of the grader. Stage~1 executes both \nothink{} drafts against the same reference tests that later score the answer, so an accepted code query is correct by construction and code accept precision is 100\% by definition rather than by measurement. To bound what this hides, we re-gated the stored HumanEval drafts on half of each problem's assertions and scored the returned program against the full suite, on the problems whose drafts the logs stored in full. Accept precision falls from a forced 100\% to 92.5\% at 8B, 97.9\% at 14B and 98.1\% at 32B, the band of the 90.8\% we report on MATH-500, and accuracy falls by at most 3.7 points. The code gains in Table~\ref{tab:main} are therefore gains under a gate that reads the evaluation tests.

\paragraph{Evaluation scope.}
The main results cover Qwen3 models on math and code together with DeepSeek-V3.2 on math, so behavior on other open-ended task families remains untested. Accuracy parity also depends on model capability, and Section~\ref{sec:analysis} reports that \ours{} trails AT on MATH-500 at the 0.6B--4B scales. AIME results in Appendix~\ref{app:additional_results} are exploratory rather than confirmatory benchmark coverage.

\paragraph{Deployment assumptions.}
Stage~1 requires only generated text and is compatible with text-only APIs, whereas the entropy-budgeted Stage~2 additionally assumes access to token-level uncertainty signals and sufficient control over thinking budgets or stop behavior. Stage~2 also fits its entropy-budget map on a probe run that overlaps the math evaluation set, so the reported budgets are not a held-out estimate and new task families require refitting. Hosted APIs exposing only a subset of these controls can still run Stage~1 routing without the full entropy-budgeted gains. Table~\ref{tab:routing} and the MATH-500 row of Table~\ref{tab:think_budget} report total-token accounting over drafts, thinking, and answer tokens, but realized cost depends on provider pricing. Wall-clock latency is measured on a single A100-80GB configuration with parallel \nothink{} drafts, so absolute speedups may vary with serving stack, batching, API latency, and hardware.

\section*{Acknowledgements}
This research was supported by Basic Science Research Program through the National Research Foundation of Korea(NRF) funded by the Ministry of Education(NRF-2021R1A6A1A03045425). This work was supported by Institute for Information \& communications Technology Promotion(IITP) grant funded by the Korea government(MSIT). (RS-2024-00398115, Research on the reliability and coherence of outcomes produced by Generative AI). This research was supported by Culture, Sports and Tourism R\&D Program, funded by the Ministry of Culture, Sports and Tourism, through the Korea Culture Technology Planning and Evaluation Institute, an affiliated institute of the Korea Creative Content Agency in 2026 (Project Name: Development of an AI Agent Integrating Korean Language Knowledge for Personalized Language Consultation Services, Project Number: RS-2026-25506607, Contribution Rate: 25\%). This work was supported by Institute of Information \& communications Technology Planning \& Evaluation (IITP) under the artificial intelligence star fellowship support program to nurture the best talents (IITP-2026-RS-2025-02304828) grant funded by the Korea government(MSIT).

\bibliography{references}

@inproceedings{adaptthink2025,
    title = "{A}dapt{T}hink: Reasoning Models Can Learn When to Think",
    author = "Zhang, Jiajie  and
      Lin, Nianyi  and
      Hou, Lei  and
      Feng, Ling  and
      Li, Juanzi",
    editor = "Christodoulopoulos, Christos  and
      Chakraborty, Tanmoy  and
      Rose, Carolyn  and
      Peng, Violet",
    booktitle = "Proceedings of the 2025 Conference on Empirical Methods in Natural Language Processing",
    month = nov,
    year = "2025",
    address = "Suzhou, China",
    publisher = "Association for Computational Linguistics",
    url = "https://aclanthology.org/2025.emnlp-main.184/",
    doi = "10.18653/v1/2025.emnlp-main.184",
    pages = "3716--3730",
    ISBN = "979-8-89176-332-6"
}

@inproceedings{thinkswitcher2025,
    title = "{T}hink{S}witcher: When to Think Hard, When to Think Fast",
    author = "Liang, Guosheng  and
      Zhong, Longguang  and
      Yang, Ziyi  and
      Quan, Xiaojun",
    editor = "Christodoulopoulos, Christos  and
      Chakraborty, Tanmoy  and
      Rose, Carolyn  and
      Peng, Violet",
    booktitle = "Findings of the Association for Computational Linguistics: EMNLP 2025",
    month = nov,
    year = "2025",
    address = "Suzhou, China",
    publisher = "Association for Computational Linguistics",
    url = "https://aclanthology.org/2025.findings-emnlp.278/",
    doi = "10.18653/v1/2025.findings-emnlp.278",
    pages = "5185--5201",
    ISBN = "979-8-89176-335-7"
}

@inproceedings{asrr2025,
    title = "When to Continue Thinking: Adaptive Thinking Mode Switching for Efficient Reasoning",
    author = "Zhang, Xiaoyun  and
      Ruan, Jingqing  and
      Ma, Xing  and
      Zhu, Yawen  and
      Zhao, Haodong  and
      Li, Hao  and
      Chen, Jiansong  and
      Zeng, Ke  and
      Cai, Xunliang",
    editor = "Christodoulopoulos, Christos  and
      Chakraborty, Tanmoy  and
      Rose, Carolyn  and
      Peng, Violet",
    booktitle = "Findings of the Association for Computational Linguistics: EMNLP 2025",
    month = nov,
    year = "2025",
    address = "Suzhou, China",
    publisher = "Association for Computational Linguistics",
    url = "https://aclanthology.org/2025.findings-emnlp.310/",
    doi = "10.18653/v1/2025.findings-emnlp.310",
    pages = "5808--5828",
    ISBN = "979-8-89176-335-7"
}

@article{selfbudgeter2025,
  title={SelfBudgeter: Adaptive Token Allocation for Efficient LLM Reasoning},
  author={Li, Zheng and Dong, Qingxiu and Ma, Jingyuan and Zhang, Di and Jia, Kai and Sui, Zhifang},
  journal={arXiv preprint arXiv:2505.11274},
  year={2025},
  eprint={2505.11274},
  archivePrefix={arXiv},
  primaryClass={cs.AI},
  url={https://arxiv.org/abs/2505.11274}
}

@misc{hbpo2025,
      title={Hierarchical Budget Policy Optimization for Adaptive Reasoning}, 
      author={Shangke Lyu and Linjuan Wu and Yuchen Yan and Xingyu Wu and Hao Li and Yongliang Shen and Peisheng Jiang and Weiming Lu and Jun Xiao and Yueting Zhuang},
      year={2025},
      eprint={2507.15844},
      archivePrefix={arXiv},
      primaryClass={cs.AI},
      url={https://arxiv.org/abs/2507.15844}, 
}

@article{routetoreason2025,
  title={Route to Reason: Adaptive Routing for LLM and Reasoning Strategy Selection},
  author={Pan, Zhihong and Zhang, Kai and Zhao, Yuze and Han, Yupeng},
  journal={arXiv preprint arXiv:2505.19435},
  year={2025},
  eprint={2505.19435},
  archivePrefix={arXiv},
  primaryClass={cs.CL},
  url={https://arxiv.org/abs/2505.19435}
}

@misc{demystifying2025,
      title={Demystifying Hybrid Thinking: Can LLMs Truly Switch Between Think and No-Think?}, 
      author={Shouren Wang and Wang Yang and Xianxuan Long and Qifan Wang and Vipin Chaudhary and Xiaotian Han},
      year={2025},
      eprint={2510.12680},
      archivePrefix={arXiv},
      primaryClass={cs.LG},
      url={https://arxiv.org/abs/2510.12680}, 
}

@article{reasoningbudget2025,
  title={Reasoning on a Budget: A Survey of Adaptive and Controllable Test-Time Compute in LLMs},
  author={Alomrani, Mohammad Ali and Zhang, Yingxue and Li, Derek and Sun, Qianyi and Pal, Soumyasundar and Zhang, Zhanguang and Hu, Yaochen and Ajwani, Rohan Deepak and Valkanas, Antonios and Karimi, Raika and Cheng, Peng and Wang, Yunzhou and Liao, Pengyi and Huang, Hanrui and Wang, Bin and Hao, Jianye and Coates, Mark},
  journal={arXiv preprint arXiv:2507.02076},
  year={2025},
  eprint={2507.02076},
  archivePrefix={arXiv},
  primaryClass={cs.AI},
  url={https://arxiv.org/abs/2507.02076}
}

@article{qwen3,
  title={Qwen3 Technical Report},
  author={{Qwen Team}},
  journal={arXiv preprint arXiv:2505.09388},
  year={2025}
}

@misc{gemma4,
  title={{Gemma 4} Model Card},
  author={{Google DeepMind}},
  year={2026},
  howpublished={Google AI for Developers},
  url={https://ai.google.dev/gemma/docs/core/model_card_4}
}

@article{
stopoverthinking2025,
title={Stop Overthinking: A Survey on Efficient Reasoning for Large Language Models},
author={Yang Sui and Yu-Neng Chuang and Guanchu Wang and Jiamu Zhang and Tianyi Zhang and Jiayi Yuan and Hongyi Liu and Andrew Wen and Shaochen Zhong and Na Zou and Hanjie Chen and Xia Hu},
journal={Transactions on Machine Learning Research},
issn={2835-8856},
year={2025},
url={https://openreview.net/forum?id=HvoG8SxggZ},
note={}
}

@article{
efficientreasoning2025,
title={Efficient Reasoning Models: A Survey},
author={Sicheng Feng and Gongfan Fang and Xinyin Ma and Xinchao Wang},
journal={Transactions on Machine Learning Research},
issn={2835-8856},
year={2025},
url={https://openreview.net/forum?id=sySqlxj8EB},
note={}
}

@inproceedings{
selfconsistency2023,
title={Self-Consistency Improves Chain of Thought Reasoning in Language Models},
author={Xuezhi Wang and Jason Wei and Dale Schuurmans and Quoc V Le and Ed H. Chi and Sharan Narang and Aakanksha Chowdhery and Denny Zhou},
booktitle={The Eleventh International Conference on Learning Representations },
year={2023},
url={https://openreview.net/forum?id=1PL1NIMMrw}
}

@article{deepseekv32,
  title={DeepSeek-V3.2: Pushing the Frontier of Open Large Language Models},
  author={{DeepSeek-AI} and Liu, Aixin and Mei, Aoxue and Lin, Bangcai and Xue, Bing and Wang, Bingxuan and Xu, Bingzheng and Wu, Bochao and Zhang, Bowei and Lin, Chaofan and others},
  journal={arXiv preprint arXiv:2512.02556},
  year={2025},
  eprint={2512.02556},
  archivePrefix={arXiv},
  primaryClass={cs.CL},
  url={https://arxiv.org/abs/2512.02556}
}

@article{deepseekr1,
  title={DeepSeek-R1: Incentivizing Reasoning Capability in LLMs via Reinforcement Learning},
  author={DeepSeek-AI},
  journal={arXiv preprint arXiv:2501.12948},
  year={2025}
}

@inproceedings{olympiadbench,
    title = "{O}lympiad{B}ench: A Challenging Benchmark for Promoting {AGI} with Olympiad-Level Bilingual Multimodal Scientific Problems",
    author = "He, Chaoqun  and
      Luo, Renjie  and
      Bai, Yuzhuo  and
      Hu, Shengding  and
      Thai, Zhen  and
      Shen, Junhao  and
      Hu, Jinyi  and
      Han, Xu  and
      Huang, Yujie  and
      Zhang, Yuxiang  and
      Liu, Jie  and
      Qi, Lei  and
      Liu, Zhiyuan  and
      Sun, Maosong",
    editor = "Ku, Lun-Wei  and
      Martins, Andre  and
      Srikumar, Vivek",
    booktitle = "Proceedings of the 62nd Annual Meeting of the Association for Computational Linguistics (Volume 1: Long Papers)",
    month = aug,
    year = "2024",
    address = "Bangkok, Thailand",
    publisher = "Association for Computational Linguistics",
    url = "https://aclanthology.org/2024.acl-long.211/",
    doi = "10.18653/v1/2024.acl-long.211",
    pages = "3828--3850"
}

@inproceedings{
entropycalibration2023,
title={Semantic Uncertainty: Linguistic Invariances for Uncertainty Estimation in Natural Language Generation},
author={Lorenz Kuhn and Yarin Gal and Sebastian Farquhar},
booktitle={The Eleventh International Conference on Learning Representations },
year={2023},
url={https://openreview.net/forum?id=VD-AYtP0dve}
}

@inproceedings{
math500,
title={Measuring Mathematical Problem Solving With the {MATH} Dataset},
author={Dan Hendrycks and Collin Burns and Saurav Kadavath and Akul Arora and Steven Basart and Eric Tang and Dawn Song and Jacob Steinhardt},
booktitle={Thirty-fifth Conference on Neural Information Processing Systems Datasets and Benchmarks Track (Round 2)},
year={2021},
url={https://openreview.net/forum?id=7Bywt2mQsCe}
}

@inproceedings{
lightman2023lets,
title={Let's Verify Step by Step},
author={Hunter Lightman and Vineet Kosaraju and Yuri Burda and Harrison Edwards and Bowen Baker and Teddy Lee and Jan Leike and John Schulman and Ilya Sutskever and Karl Cobbe},
booktitle={The Twelfth International Conference on Learning Representations},
year={2024},
url={https://openreview.net/forum?id=v8L0pN6EOi}
}

@article{arc,
  title={Think you have Solved Question Answering? Try {ARC}, the {AI2} Reasoning Challenge},
  author={Clark, Peter and Cowhey, Isaac and Etzioni, Oren and Khot, Tushar and Sabharwal, Ashish and Schoenick, Carissa and Tafjord, Oyvind},
  journal={arXiv preprint arXiv:1803.05457},
  year={2018}
}

@inproceedings{
gpqa,
title={{GPQA}: A Graduate-Level Google-Proof Q\&A Benchmark},
author={David Rein and Betty Li Hou and Asa Cooper Stickland and Jackson Petty and Richard Yuanzhe Pang and Julien Dirani and Julian Michael and Samuel R. Bowman},
booktitle={First Conference on Language Modeling},
year={2024},
url={https://openreview.net/forum?id=Ti67584b98}
}

@inproceedings{mmlupro,
author = {Wang, Yubo and Ma, Xueguang and Zhang, Ge and Ni, Yuansheng and Chandra, Abhranil and Guo, Shiguang and Ren, Weiming and Arulraj, Aaran and He, Xuan and Jiang, Ziyan and Li, Tianle and Ku, Max and Wang, Kai and Zhuang, Alex and Fan, Rongqi and Yue, Xiang and Chen, Wenhu},
title = {MMLU-Pro: a more robust and challenging multi-task language understanding benchmark},
year = {2024},
isbn = {9798331314385},
publisher = {Curran Associates Inc.},
address = {Red Hook, NY, USA},
booktitle = {Proceedings of the 38th International Conference on Neural Information Processing Systems},
articleno = {3018},
numpages = {25},
location = {Vancouver, BC, Canada},
series = {NIPS '24}
}

@article{humaneval,
  title={Evaluating Large Language Models Trained on Code},
  author={Chen, Mark and Tworek, Jerry and Jun, Heewoo and Yuan, Qiming and de Oliveira Pinto, Henrique Ponde and Kaplan, Jared and Edwards, Harri and Burda, Yuri and Joseph, Nicholas and Brockman, Greg and others},
  journal={arXiv preprint arXiv:2107.03374},
  year={2021}
}

@article{mbpp,
  title={Program Synthesis with Large Language Models},
  author={Austin, Jacob and Odena, Augustus and Nye, Maxwell and Bosma, Maarten and Michalewski, Henryk and Dohan, David and Jiang, Ellen and Cai, Carrie and Terry, Michael and Le, Quoc and Sutton, Charles},
  journal={arXiv preprint arXiv:2108.07732},
  year={2021},
  eprint={2108.07732},
  archivePrefix={arXiv},
  primaryClass={cs.PL},
  url={https://arxiv.org/abs/2108.07732}
}

@misc{aime2024,
  title={{AIME} 2024},
  author={Jia, Maxwell},
  year={2024},
  howpublished={Hugging Face dataset},
  url={https://huggingface.co/datasets/Maxwell-Jia/AIME_2024}
}

@misc{aime2025,
  title={AIME 2025 - Unified Test-Time Scaling Format},
  author={Test-Time Compute Organization},
  year={2025},
  publisher={Hugging Face},
  howpublished={\url{https://huggingface.co/datasets/test-time-compute/aime_2025}}
}

@article{qwen25math,
  title={Qwen2.5-Math Technical Report: Toward Mathematical Expert Model via Self-Improvement},
  author={Yang, An and Zhang, Beichen and Hui, Binyuan and Gao, Bofei and Yu, Bowen and Li, Chengpeng and Liu, Dayiheng and Tu, Jianhong and Zhou, Jingren and Lin, Junyang and others},
  journal={arXiv preprint arXiv:2409.12122},
  year={2024},
  eprint={2409.12122},
  archivePrefix={arXiv},
  primaryClass={cs.CL},
  url={https://arxiv.org/abs/2409.12122}
}

@article{thinkingtrap2025,
  title={Do Thinking Tokens Help or Trap? Towards More Efficient Large Reasoning Model},
  author={Ding, Bowen and others},
  journal={arXiv preprint arXiv:2506.23840},
  year={2025}
}

@inproceedings{vllm,
author = {Kwon, Woosuk and Li, Zhuohan and Zhuang, Siyuan and Sheng, Ying and Zheng, Lianmin and Yu, Cody Hao and Gonzalez, Joseph and Zhang, Hao and Stoica, Ion},
title = {Efficient Memory Management for Large Language Model Serving with PagedAttention},
year = {2023},
isbn = {9798400702297},
publisher = {Association for Computing Machinery},
address = {New York, NY, USA},
url = {https://doi.org/10.1145/3600006.3613165},
doi = {10.1145/3600006.3613165},
booktitle = {Proceedings of the 29th Symposium on Operating Systems Principles},
pages = {611–626},
numpages = {16},
location = {Koblenz, Germany},
series = {SOSP '23}
}

@article{
frugalgpt2023,
title={Frugal{GPT}: How to Use Large Language Models While Reducing Cost and Improving Performance},
author={Lingjiao Chen and Matei Zaharia and James Zou},
journal={Transactions on Machine Learning Research},
issn={2835-8856},
year={2024},
url={https://openreview.net/forum?id=cSimKw5p6R},
note={Featured Certification}
}

@inproceedings{
routellm2024,
title={Route{LLM}: Learning to Route {LLM}s from Preference Data},
author={Isaac Ong and Amjad Almahairi and Vincent Wu and Wei-Lin Chiang and Tianhao Wu and Joseph E. Gonzalez and M Waleed Kadous and Ion Stoica},
booktitle={The Thirteenth International Conference on Learning Representations},
year={2025},
url={https://openreview.net/forum?id=8sSqNntaMr}
}

@inproceedings{
snell2024scaling,
title={Scaling {LLM} Test-Time Compute Optimally Can be More Effective than Scaling Parameters for Reasoning},
author={Charlie Victor Snell and Jaehoon Lee and Kelvin Xu and Aviral Kumar},
booktitle={The Thirteenth International Conference on Learning Representations},
year={2025},
url={https://openreview.net/forum?id=4FWAwZtd2n}
}

@inproceedings{
universalsc2024,
title={Universal Self-Consistency for Large Language Models},
author={Xinyun Chen and Renat Aksitov and Uri Alon and Jie Ren and Kefan Xiao and Pengcheng Yin and Sushant Prakash and Charles Sutton and Xuezhi Wang and Denny Zhou},
booktitle={ICML 2024 Workshop on In-Context Learning},
year={2024},
url={https://openreview.net/forum?id=LjsjHF7nAN}
}

@misc{graves2016act,
      title={Adaptive Computation Time for Recurrent Neural Networks}, 
      author={Alex Graves},
      year={2017},
      eprint={1603.08983},
      archivePrefix={arXiv},
      primaryClass={cs.NE},
      url={https://arxiv.org/abs/1603.08983}, 
}

@inproceedings{
schuster2022calm,
title={Confident Adaptive Language Modeling},
author={Tal Schuster and Adam Fisch and Jai Gupta and Mostafa Dehghani and Dara Bahri and Vinh Q. Tran and Yi Tay and Donald Metzler},
booktitle={Advances in Neural Information Processing Systems},
editor={Alice H. Oh and Alekh Agarwal and Danielle Belgrave and Kyunghyun Cho},
year={2022},
url={https://openreview.net/forum?id=uLYc4L3C81A}
}

@article{kadavath2022know,
  title={Language Models (Mostly) Know What They Know},
  author={Kadavath, Saurav and Conerly, Tom and Askell, Amanda and Henighan, Tom and Drain, Dawn and Perez, Ethan and Schiefer, Nicholas and Hatfield-Dodds, Zac and DasSarma, Nova and Tran-Johnson, Eli and others},
  journal={arXiv preprint arXiv:2207.05221},
  year={2022}
}

@article{ayer1955empirical,
  title={An Empirical Distribution Function for Sampling with Incomplete Information},
  author={Ayer, Miriam and Brunk, H. D. and Ewing, G. M. and Reid, W. T. and Silverman, Edward},
  journal={The Annals of Mathematical Statistics},
  volume={26},
  number={4},
  pages={641--647},
  year={1955},
  doi={10.1214/aoms/1177728423}
}

@inproceedings{leviathan2023speculativedecoding,
  title={Fast inference from transformers via speculative decoding},
  author={Leviathan, Yaniv and Kalman, Matan and Matias, Yossi},
  booktitle={International Conference on Machine Learning},
  pages={19274--19286},
  year={2023},
  organization={PMLR}
}

@inproceedings{aggarwal2023adaptiveconsistency,
    title = "Let{'}s Sample Step by Step: Adaptive-Consistency for Efficient Reasoning and Coding with {LLM}s",
    author = "Aggarwal, Pranjal  and
      Madaan, Aman  and
      Yang, Yiming  and
      Mausam",
    editor = "Bouamor, Houda  and
      Pino, Juan  and
      Bali, Kalika",
    booktitle = "Proceedings of the 2023 Conference on Empirical Methods in Natural Language Processing",
    month = dec,
    year = "2023",
    address = "Singapore",
    publisher = "Association for Computational Linguistics",
    url = "https://aclanthology.org/2023.emnlp-main.761/",
    doi = "10.18653/v1/2023.emnlp-main.761",
    pages = "12375--12396"
}

@misc{gpt55,
  title={Introducing {GPT-5.5}},
  author={{OpenAI}},
  year={2026},
  howpublished={OpenAI product release},
  url={https://openai.com/index/introducing-gpt-5-5/}
}

@misc{claudethinking,
  title={Building with Extended Thinking},
  author={{Anthropic}},
  year={2026},
  howpublished={Claude API documentation},
  url={https://platform.claude.com/docs/en/build-with-claude/extended-thinking}
}

@misc{yu2025minicpmv45cookingefficient,
      title={MiniCPM-V 4.5: Cooking Efficient MLLMs via Architecture, Data, and Training Recipe},
      author={Yu, Tianyu and Wang, Zefan and Wang, Chongyi and Huang, Fuwei and Ma, Wenshuo and He, Zhihui and Cai, Tianchi and Chen, Weize and Huang, Yuxiang and Zhao, Yuanqian and others},
      year={2025},
      eprint={2509.18154},
      archivePrefix={arXiv},
      primaryClass={cs.CV},
      url={https://arxiv.org/abs/2509.18154},
}

@misc{nvidia2026nemotron3nanoomni,
      title={Nemotron 3 Nano Omni: Efficient and Open Multimodal Intelligence},
      author={NVIDIA},
      year={2026},
      eprint={2604.24954},
      archivePrefix={arXiv},
      primaryClass={cs.LG},
      url={https://arxiv.org/abs/2604.24954},
}

@article{ma2025nothinking,
  title={Reasoning Models Can Be Effective Without Thinking},
  author={Ma, Wenjie and He, Jingxuan and Snell, Charlie and Griggs, Tyler and Min, Sewon and Zaharia, Matei},
  journal={arXiv preprint arXiv:2504.09858},
  year={2025}
}

@article{wang2024penny,
  title={Make Every Penny Count: Difficulty-Adaptive Self-Consistency for Cost-Efficient Reasoning},
  author={Wang, Xinglin and Feng, Shaoxiong and Li, Yiwei and Yuan, Peiwen and Zhang, Yueqi and Tan, Chuyi and Pan, Boyuan and Hu, Yao and Li, Kan},
  journal={arXiv preprint arXiv:2408.13457},
  year={2024}
}

@article{lee2026acoer,
  title={Beyond Penalizing Mistakes: Stabilizing Efficiency Training in Large Reasoning Models via Adaptive Correct-Only Rewards},
  author={Lee, Jungseob and Lee, Seungyoon and Hong, Seongtae and Kim, Minhyuk and Park, Chanjun and Lim, Heuiseok},
  journal={arXiv preprint arXiv:2606.22716},
  year={2026},
  eprint={2606.22716},
  archivePrefix={arXiv},
  url={https://arxiv.org/abs/2606.22716}
}

@article{lee2026pricecheck,
  title={Risk-Controlled Selective {LLM} Answering by Pricing Label-Free Checks},
  author={Lee, Dongyub Jude and Lee, Jungseob and Park, Chanjun and Moon, Hyeonseok and Lim, Heuiseok},
  journal={arXiv preprint arXiv:2609.37493},
  year={2026},
  eprint={2609.37493},
  archivePrefix={arXiv},
  url={https://arxiv.org/abs/2609.37493}
}

@article{lee2026directional,
  title={Distilling Directional Verification},
  author={Lee, Jungseob and Eo, Sugyeong and Hong, Seongtae and Lee, Seungyoon and Park, Chanjun and Seo, Jaehyung and Lim, Heuiseok},
  journal={arXiv preprint arXiv:2610.00997},
  year={2026},
  eprint={2610.00997},
  archivePrefix={arXiv},
  url={https://arxiv.org/abs/2610.00997}
}

@article{lee2026offpolicy,
  title={Recovering Off-Policy Supervision for Speculative Decoding},
  author={Lee, Jungseob and Park, Chanjun and Eo, Sugyeong and Moon, Hyeonseok},
  journal={arXiv preprint arXiv:2609.38795},
  year={2026},
  eprint={2609.38795},
  archivePrefix={arXiv},
  url={https://arxiv.org/abs/2609.38795}
}

@article{lee2026cast,
  title={{CAST}: Cost-Aware Speculative Trees from One-Pass Block Drafters},
  author={Lee, Jungseob and Eo, Sugyeong},
  journal={arXiv preprint arXiv:2610.00321},
  year={2026},
  eprint={2610.00321},
  archivePrefix={arXiv},
  url={https://arxiv.org/abs/2610.00321}
}

@article{lee2026madara,
  title={To Isolate or to Score? Model-Adaptive Assessment for Cost-Efficient Multi-Agent {RAG}},
  author={Lee, Jungseob and Park, Chanjun and Lim, Heuiseok},
  journal={arXiv preprint arXiv:2606.25191},
  year={2026},
  eprint={2606.25191},
  archivePrefix={arXiv},
  url={https://arxiv.org/abs/2606.25191}
}

@article{lee2026glance,
  title={Vision Is Not Overhead: One-Pass Block Drafting for Lossless Speculative Decoding in Vision-Language Models},
  author={Lee, Jungseob and Hong, Seongtae and Lee, Dongyub Jude and Park, Chanjun and Seo, Jaehyung and Eo, Sugyeong and Lim, Heuiseok},
  journal={arXiv preprint arXiv:2609.00355},
  year={2026},
  eprint={2609.00355},
  archivePrefix={arXiv},
  url={https://arxiv.org/abs/2609.00355}
}

@article{lee2026redline,
  title={Faster Block-Diffusion Serving with Distribution-Free Risk Guarantees},
  author={Lee, Jungseob and Lee, Dongyub Jude and Park, Chanjun and Eo, Sugyeong and Lim, Heuiseok},
  journal={arXiv preprint arXiv:2609.33887},
  year={2026},
  eprint={2609.33887},
  archivePrefix={arXiv},
  url={https://arxiv.org/abs/2609.33887}
}

\clearpage
\appendix
\suppressfloats[t]

\section{Additional Cross-Model and Small-Sample Results}
\label{app:additional_results}

\paragraph{AIME 2024/2025.} Table~\ref{tab:aime_smallsample} reports \ours{} on AIME 2024 and 2025 for Qwen3-8B and DeepSeek-V3.2. We report these as exploratory evidence rather than confirmatory comparison. On Qwen3-8B, \ours{} gains $+$6.7 on AIME 2024 and $+$3.4 on AIME 2025 over AT, with 32\% and 29\% thinking-token reductions. On DeepSeek-V3.2, \ours{} matches AT on AIME 2024 and exceeds AT by 10 points, at 42\% and 28\% fewer thinking tokens.

\begin{table}[t]
\centering
\small
\setlength{\tabcolsep}{4pt}
\resizebox{\columnwidth}{!}{%
\begin{tabular}{@{}ll rrr r@{}}
\toprule
\textbf{Model} & \textbf{Benchmark} & \textbf{NT} & \textbf{AT} & \textbf{\ours{}} \footnotesize(vs AT) & \textbf{Think\,$\downarrow$} \\
\midrule
\multirow{2}{*}{Qwen3-8B}
 & AIME 2024 & 26.7 & 60.0 & \textbf{66.7}\,{\color{posdelta}\footnotesize($+$6.7)} & 32\% \\
 & AIME 2025 & 26.7 & 53.3 & \textbf{56.7}\,{\color{posdelta}\footnotesize($+$3.4)} & 29\% \\
\midrule
\multirow{2}{*}{DeepSeek-V3.2}
 & AIME 2024 & 66.7 & 70.0 & 70.0\,{\footnotesize($+$0.0)} & 42\% \\
 & AIME 2025 & 53.3 & 40.0 & \textbf{50.0}\,{\color{posdelta}\footnotesize($+$10.0)} & 28\% \\
\bottomrule
\end{tabular}%
}
\caption{\ours{} on AIME 2024 / 2025 for Qwen3-8B and DeepSeek-V3.2. Exploratory results.}
\label{tab:aime_smallsample}

\end{table}

\section{Truncation Analysis and Budget Sweep}
\label{app:truncation}

The one-phase AT accuracies reported in this appendix (Tables~\ref{tab:truncation} and~\ref{tab:olymp_truncation}) use a diagnostic evaluation protocol designed for truncated one-phase outputs rather than the main-table protocol. Numbers are therefore not directly comparable to Table~\ref{tab:iso_accuracy}'s \ours{} accuracy. The trend, namely truncation at low budget and a plateau at high budget, is stable across evaluation protocols.

\begin{table}[t]
\centering
\small
\setlength{\tabcolsep}{4pt}
\begin{tabular}{@{}r rr r rr@{}}
\toprule
\textbf{B} & \textbf{Acc} & \textbf{Trunc\%} & \textbf{B} & \textbf{Acc} & \textbf{Trunc\%} \\
\midrule
2K   & 13.5 & 86.0 & 22K & 71.8 & 1.8 \\
4K   & 40.0 & 49.2 & 24K & 70.8 & 2.8 \\
6K   & 55.0 & 28.7 & 26K & 71.8 & 1.5 \\
7K   & 59.0 & 23.5 & 28K & 71.8 & 0.2 \\
8K   & 61.0 & 19.5 & 30K & 72.2 & 0.2 \\
9K   & 65.7 & 14.5 & 32K & 72.0 & 0.8 \\
10K  & 66.0 & 12.5 & 36K & 71.8 & 0.8 \\
11K  & 67.5 & 10.5 & 40K & 72.0 & 1.0 \\
12K  & 68.8 &  9.2 & 44K & 71.2 & 0.5 \\
13K  & 68.8 &  6.8 & 48K & \textbf{72.2} & 0.5 \\
14K  & 69.8 &  5.5 & 52K & 71.5 & 0.8 \\
16K  & 69.5 &  5.0 & 56K & 71.8 & 0.2 \\
18K  & 70.0 &  3.8 & 60K & 71.8 & 0.5 \\
20K  & 71.5 &  2.2 & & & \\
\midrule
\multicolumn{6}{@{}l}{\textit{\ours{} (two-stage, 1{,}743 mean thinking tokens): \textbf{88.2}}} \\
\bottomrule
\end{tabular}
\caption{One-phase AT accuracy under token budget on MATH-500 (Qwen3-8B).}
\label{tab:truncation}
\end{table}

\begin{table}[t]
\centering
\small
\setlength{\tabcolsep}{4pt}
\begin{tabular}{@{}r rr@{}}
\toprule
\textbf{Budget} & \textbf{Acc} & \textbf{Trunc\%} \\
\midrule
 2K &  1.9 & 98.7 \\
 4K & 14.6 & 79.2 \\
 8K & 36.7 & 45.8 \\
16K & 53.8 & 16.5 \\
32K & 58.9 &  2.3 \\
49K & 58.1 &  3.8 \\
\bottomrule
\end{tabular}
\caption{One-phase AT on OlympiadBench (Qwen3-8B).}
\label{tab:olymp_truncation}
\end{table}

\section{\texorpdfstring{Hyperparameter Ablation ($K$)}{Hyperparameter Ablation (K)} and Multiple-Choice Scope}
\label{app:hparam_ablation}

\paragraph{Full $K$ sweep at strict budget (MATH-500).}
Table~\ref{tab:k_full_sweep} reports a sweep over $K \in \{1, 2, 3, 4\}$ on the first 200 MATH-500 problems, with every routed query capped at 4{,}096 thinking tokens. This is a separate run from the main experiments, which use all 500 problems at this scale and an entropy-predicted budget, and it redraws its drafts independently, so its accept rates and accuracies are comparable across $K$ but not against Table~\ref{tab:main}. Accuracy peaks at $K{=}3$ (89.5\%), with $K{=}2$ close behind (89.0\%). $K{=}1$ degenerates to always-NT routing and underperforms (81.0\%), while $K{=}4$ slightly regresses (88.5\%) despite consuming 22\% more thinking tokens than $K{=}3$. We therefore use $K{=}2$ as the practical efficiency point. Moving from $K{=}2$ to $K{=}3$ adds only a 0.5-point accuracy gain while increasing average thinking tokens from 772 to 939, and $K{=}4$ uses more thinking tokens without further accuracy. We adopt $K{=}2$ as the default in all other tables.

\paragraph{Multiple-choice task scope.}
The AT and NT columns of Table~\ref{tab:mcq} separate two effects that Section~\ref{sec:analysis} does not distinguish. \think{} mode itself remains useful on all three benchmarks, with AT above NT at 92.3 against 86.1 on ARC-Challenge, 37.7 against 17.8 on MMLU-Pro, and 56.6 against 47.0 on GPQA-D. What collapses is the unanimity signal that Stage~1 reads. Its association with AT correctness is not significant on MMLU-Pro ($p = 0.49$) or GPQA-D ($p = 0.90$), and on ARC-Challenge it reaches significance only at a magnitude of 0.13. These runs keep the $K{=}2$ draft procedure of the main experiments and substitute an option-letter equivalence check for the math and code equivalence functions of \S\ref{app:equivalence}. Together with the collision-probability argument given there, the near-zero association is why multiple-choice tasks stay outside the routing scope.

\begin{table}[t]
\centering
\small
\begin{tabular}{@{}lccc@{}}
\toprule
$K$ & Accuracy (\%) & NT-route rate & Avg.\ think tok.\ \\
\midrule
1 & 81.0 & 1.000 & 0 \\
2 & 89.0 & 0.810 & 772 \\
\textbf{3} & \textbf{89.5} & 0.760 & 939 \\
4 & 88.5 & 0.705 & 1{,}149 \\
\bottomrule
\end{tabular}
\caption{Accuracy and routing behavior as a function of $K$ on Qwen3-8B over the first 200 MATH-500 problems, with a strict 4{,}096-token budget cap for each query. Separate run from Table~\ref{tab:main}.}
\label{tab:k_full_sweep}
\end{table}

\begin{table}[t]
\centering
\small
\begin{tabular}{@{}lcccc@{}}
\toprule
Benchmark & AT & NT & $r_{pb}$ & $p$ \\
\midrule
ARC-C     & \textbf{92.3} & 86.1 & 0.13 & $<$0.001 \\
MMLU-Pro  & \textbf{37.7} & 17.8 & 0.06 & 0.49 \\
GPQA-D    & \textbf{56.6} & 47.0 & $-$0.009 & 0.90 \\
\bottomrule
\end{tabular}
\caption{Multiple-choice benchmarks on Qwen3-8B. $r_{pb}$ is the point-biserial correlation between draft unanimity and AT correctness.}
\label{tab:mcq}

\end{table}

\section{Budget Ablation}
\label{app:budget}

Tables~\ref{tab:budget} and~\ref{tab:temp} sweep the thinking-budget cap for each query and the draft sampling temperature. Accuracy plateaus once the budget exceeds the entropy-predicted average (about 2.8K on MATH-500), with no benefit from larger caps. Below the predicted average, accuracy degrades sharply, supporting the entropy-to-budget mapping in Eq.~\ref{eq:budget_pred} and Figure~\ref{fig:entropy_budget}. The last two rows of Table~\ref{tab:budget} instead isolate Stage~2. Both serve the same Stage~1 accepts, and the control reassigns the predicted budgets of the 110 queries Stage~1 routes to \think{} mode by resampling those same budgets with replacement, so the budget distribution is matched and only the pairing between draft entropy and query is destroyed. We report its mean and standard deviation over 100 seeds. The entropy-conditioned schedule covers the realized thinking-token demand of all 110 routed queries, while the shuffled schedule under-budgets part of the set.

\begin{table}[t]
\centering
\small
\begin{tabular}{@{}r r r r@{}}
\toprule
\textbf{B} & \textbf{Acc} & \textbf{B} & \textbf{Acc} \\
\midrule
1K & 69.25 & 12K & 71.75 \\
2K & 69.75 & 14K & \textbf{73.00} \\
3K & 70.75 & 16K & 72.25 \\
4K & 70.00 & 20K & 70.00 \\
5K & 70.50 & 24K & 72.25 \\
6K & 71.25 & 28K & 71.50 \\
8K & 70.00 & 32K & 72.25 \\
10K & 71.00 & 49K & 71.25 \\
\midrule
\multicolumn{4}{@{}l}{\textit{Strategy comparison under the main evaluation protocol}} \\
\multicolumn{2}{@{}l}{Shuffled budget (100 seeds)} & \multicolumn{2}{r}{$87.3 \pm 0.3$} \\
\multicolumn{2}{@{}l}{\ours{} (entropy-conditioned)} & \multicolumn{2}{r}{\textbf{88.2}} \\
\bottomrule
\end{tabular}
\caption{Budget ablation on MATH-500 (Qwen3-8B), string-extract grader on 400 problems. The last two rows use the main evaluation protocol on all 500 problems and are not comparable to the sweep above them.}
\label{tab:budget}

\end{table}

\paragraph{Effect of $T$ (sampling temperature).}
Table~\ref{tab:temp} varies the draft sampling temperature on DeepSeek-V3.2 over a 100-problem MATH-500 subset, so its absolute values are not comparable to the DeepSeek-V3.2 row of Table~\ref{tab:main} and only the ordering across $T$ is meaningful. NT\% and SC\% are the accuracies of fixed \nothink{} generation and of Stage~1 routing, while Accept\%, Prec.\%, and $r_{pb}$ are the unanimous-accept rate, the accept precision, and the point-biserial correlation defined in Section~\ref{sec:setup}. The accept rate is lowest at $T{=}0.8$ (85.0), which is also where accept precision is highest (100.0), so a higher draft temperature trades accepted queries for precision on the queries Stage~1 still keeps. Each column peaks at a different temperature, with SC\% highest at $T{=}0.7$ (96.0), Prec.\% and $r_{pb}$ at $T{=}0.8$, and NT\% at $T{=}0.9$ (93.0), while $r_{pb}$ stays above 0.69 at every setting, so the agreement signal Stage~1 depends on survives the whole sweep. We keep $T{=}0.7$, the hosted-API default used for these runs.

\begin{table}[t]
\centering
\small
\begin{tabular}{@{}cccccc@{}}
\toprule
$T$ & NT\% & SC\% & Accept\% & Prec.\% & $r_{pb}$ \\
\midrule
0.5 & 91.0 & 95.0 & 88.0 & 98.9 & 0.744 \\
0.6 & 91.0 & 92.0 & 89.0 & 98.9 & 0.783 \\
0.7 & 89.0 & \textbf{96.0} & 90.0 & 97.8 & 0.842 \\
0.8 & 86.0 & 93.0 & 85.0 & \textbf{100.0} & \textbf{0.960} \\
0.9 & 93.0 & 94.0 & 90.0 & 98.9 & 0.692 \\
\bottomrule
\end{tabular}
\caption{Temperature sensitivity on DeepSeek-V3.2 MATH-500. $T{=}0.7$ is the default.}
\label{tab:temp}

\end{table}

\section{Implementation Details}
\label{app:implementation}

\paragraph{Qwen3-8B.}
Local serving uses vLLM~\citep{vllm}; the installed version is 0.16.0 with reasoning-output support enabled.
Drafts use $T{=}0.6$, $\text{top-}p{=}0.95$, \texttt{enable\_thinking=false}, and a maximum of 1024 tokens following Qwen3 chat-template defaults.
Escalation uses $T{=}0.0$ (greedy), \texttt{enable\_thinking=true}, and a 16{,}384-token thinking maximum, the same cap the AT baseline receives. The 4{,}096-token cap appears only in the strict-budget sweep of \S\ref{app:hparam_ablation}.
Hardware is NVIDIA A100-80GB.

\paragraph{DeepSeek-V3.2.}
We queried the hosted DeepSeek API before the 2026-04-24 alias transition, when \texttt{deepseek-reasoner} and \texttt{deepseek-chat} mapped to DeepSeek-V3.2 thinking and non-thinking modes.
Drafts use the \texttt{deepseek-chat} endpoint at API defaults.

\paragraph{Artifact access and terms.}
Model, benchmark, API, and software artifacts were used only for research evaluation under the licenses or access terms stated by their original providers. We cite the original artifact creators where the artifacts are introduced, including Section~\ref{sec:setup} and Appendix~\ref{app:implementation}. We do not redistribute third-party model weights, benchmark data, hosted API outputs as a standalone dataset, or modified third-party artifacts. Hosted-model experiments were conducted through provider APIs under their service terms.

\paragraph{Answer extraction.}
Math answers are extracted from the final \texttt{\textbackslash boxed\{...\}} span on MATH-500 and OlympiadBench, code answers are taken as the generated program and executed against the reference tests described in \S\ref{app:equivalence}, and multiple-choice answers are read from the emitted option letter.

\subsection{Evaluation Protocol Details}
\label{app:eval_protocol}
Math accuracies in the main tables use the Qwen2.5-Math semantic-equivalence grader~\citep{qwen25math} (\texttt{math\_equal}) applied to full responses. Appendix budget-sweep tables (\S\ref{app:truncation}) instead use a string-extract grader for truncation diagnosis, since one-phase responses may end mid-LaTeX and confuse the semantic grader. These numbers are therefore not directly comparable to main results. Code accuracies use execution-based equivalence against each problem's reference tests, the same tests Stage~1 reads when routing, on the full HumanEval and MBPP sets for every model. Rows of Tables~\ref{tab:main} and~\ref{tab:scale} are single-run and cover the full benchmark sizes of Section~\ref{sec:setup}, except that Qwen3-32B MATH-500 covers 348 of 500 problems and OlympiadBench covers 200 of 572 problems at Qwen3-14B and Qwen3-32B. Supervised-router rows in Table~\ref{tab:routing} use 5-fold cross-validation on benchmark-matched subsets.

\subsection{Pluggable Equivalence Functions}
\label{app:equivalence}
The routing decision hinges on whether two draft answers ``agree.'' For mathematical reasoning, we normalize extracted answers by stripping LaTeX, canonicalizing numbers, and lowercasing strings. For code generation, string comparison fails because functionally identical programs rarely match character by character. We instead execute both drafts against each problem's reference test suite in a sandboxed environment and define agreement as both drafts passing all of its tests. No tests are held out, so the suite Stage~1 reads is the suite that scores the final answer, and an accepted code query passes by construction. The equivalence module is the \emph{only} domain-specific component. Extending \ours{} to a new domain requires only implementing the appropriate $\text{eq}$, and the routing mechanism remains unchanged.

\section{Entropy--Budget Calibration}
\label{app:entropy_budget_fig}

\begin{figure}[t]
\centering
\includegraphics[width=\columnwidth]{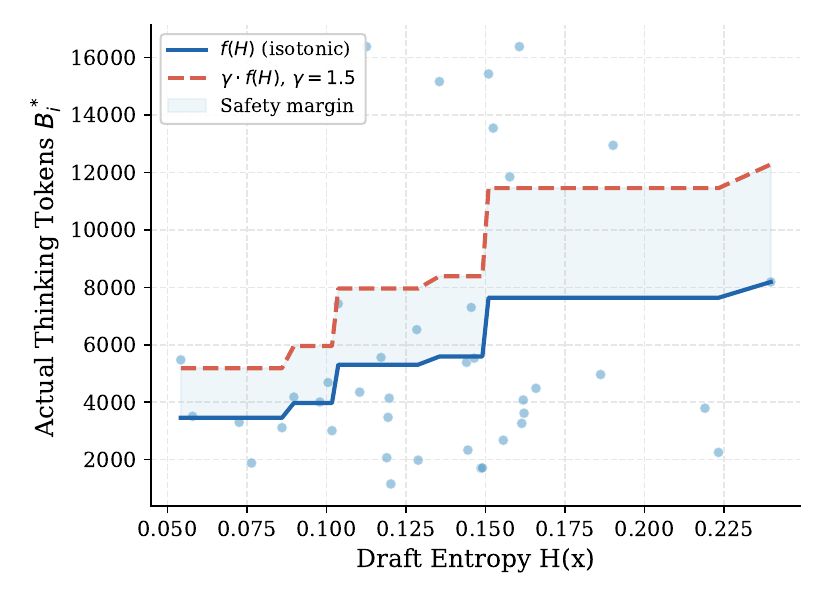}
\caption{Draft entropy vs.\ actual thinking tokens on MATH-500 calibration data. The solid line is the isotonic mapping $f(H)$. The dashed line is the safety-margin schedule $\gamma \cdot f(H)$ with $\gamma{=}1.5$.}
\label{fig:entropy_budget}
\end{figure}

Figure~\ref{fig:entropy_budget} plots the isotonic-regression mapping $f$ from draft entropy $H(q)$ to predicted thinking budget $\hat{B}(q)$ used in Stage~2 (Eq.~\ref{eq:budget_pred}). We fit the mapping label-free on a 100-problem probe run of disagreement-flagged math queries, 71 of which also appear in the evaluation set. The mapping is monotone non-decreasing by construction. Queries with diffuse draft distributions, and therefore higher entropy, receive larger thinking budgets, while low-entropy disagreement queries receive tight budgets and recover quickly. The safety margin $\gamma{=}1.5$ shifts the curve upward so the predicted budget exceeds the observed thinking-token usage at the matched entropy.

\section{Self-Verification as an Alternative Router}
\label{app:self_verify}

A natural alternative to draft unanimity is to ask the model itself whether its NT draft is correct.
We evaluated this on a MATH-500 pilot, prompting Qwen3-8B in no-think mode to judge its own draft as \emph{yes}/\emph{no}/\emph{unclear}, and routing to AT whenever the verdict was not \emph{yes}.
Using the true correctness of the draft as ground truth, verifier precision is $0.638$ (of drafts labeled \emph{yes}, $37/58$ are actually correct) and recall is $0.841$ (of correct drafts, $37/44$ are accepted).
Translated into routing behavior, the verifier \emph{detects} only $22.2\%$ of wrong NT drafts (detection rate, true-negative share among wrong drafts) while \emph{preserving} $84.1\%$ of correct ones (preservation rate).
Among the wrong drafts it does route to AT, only $3$ are actually recovered, leaving routing accuracy below \ours{}'s draft-unanimity rule.
Self-verification thus combines low sensitivity to errors with collateral loss of correct answers. We did not pursue it further in the main pipeline.

\section{Wall-clock Latency Breakdown}
\label{app:latency}

Table~\ref{tab:latency} gives the measurement behind the latency reduction reported in Section~\ref{sec:token_budget}. Timings are per-query wall clock for Qwen3-8B served locally on a single A100-80GB node under the sampling settings of \S\ref{app:implementation}, measured on MATH-500. The $K{=}2$ \nothink{} drafts are generated in parallel, so a Stage~1 accept returns after one draft pass and never enters \think{} mode. The reported \ours{} mean is the share-weighted combination of that accept path and the \think{} path, so it moves with the accept rate rather than with the thinking budget. These numbers describe one serving configuration and are not a portable speedup, since batching, serving stack, and hardware all change the absolute values.

\begin{table}[t]
\centering
\small
\begin{tabular}{@{}l rr@{}}
\toprule
\textbf{Path} & \textbf{Latency (s)} & \textbf{Share} \\
\midrule
AT (full thinking) & 67.6 & --- \\
\ours{} (overall mean) & 29.5 & 100\% \\
\ours{} (no-think exit) & 14.6 & 78\% \\
\bottomrule
\end{tabular}
\caption{Wall-clock latency on each MATH-500 query (Qwen3-8B, A100-80GB). \ours{} overall is the weighted mean across the two paths. No-think exit corresponds to Stage~1 unanimity accept.}
\label{tab:latency}
\end{table}

\end{document}